\documentclass{article} 
\usepackage[preprint]{colm2026_conference}

\usepackage{microtype}
\usepackage{hyperref}
\usepackage{url}
\usepackage{booktabs}

\usepackage{pgfplots}
\usepackage{tcolorbox}
\tcbuselibrary{breakable}
\usepackage{bm}

\usepackage{mathtools}
\usepackage{amsthm}
\theoremstyle{plain}
\newtheorem{takeaway}{Take-away}

\newtheorem{component}{Component}

\usepackage{listings}
\usepackage{amssymb}

\usepackage{multirow}
\usepackage{graphicx}
\usepackage{float}

\usepackage{wrapfig}

\definecolor{correctgreen}{rgb}{0.0, 0.5, 0.0} 
\definecolor{wrongred}{rgb}{0.8, 0.0, 0.0} 

\usepackage{CJKutf8}

\usepackage{amsmath}

\usepackage{svg}
\usepackage[T1]{fontenc}

\usepackage[utf8]{inputenc}
\usepackage{booktabs}
\usepackage{microtype}
\usepackage{url}

\usepackage{inconsolata}

\usepackage{graphicx}
\usepackage{makecell}
\usepackage{amsmath}

\usepackage{amsthm}

\usepackage{colortbl}

\usepackage{float}

\definecolor{dialoguebg}{RGB}{247,250,253}
\definecolor{analysisbg}{RGB}{250,250,250}
\definecolor{titlebg}{RGB}{235,240,246}
\definecolor{borderc}{RGB}{150,150,150}
\definecolor{turnc}{RGB}{70,70,70}
\usepackage{CJKutf8}
\usepackage{tcolorbox}
\usepackage{xcolor}

\tcbset{
  boxrule=0.65pt,
  colframe=borderc,
  arc=1.2mm,
  left=2.2mm,
  right=1.2mm,
  top=1.0mm,
  bottom=1.0mm,
  boxsep=3pt
}

\newcommand{\turnpair}[3]{
\vspace{-0.18em}
{\footnotesize\textbf{\textcolor{turnc}{Turn #1}}}\par
\noindent
\begin{minipage}[t]{0.485\textwidth}
\begin{tcolorbox}[
  colback=dialoguebg,
  title=\textbf{Dialogue},
  colbacktitle=titlebg,
  coltitle=black,
  fonttitle=\footnotesize,
  height=3.65cm,
  valign=top
]
\footnotesize
#2
\vspace*{\fill}
\end{tcolorbox}
\end{minipage}\hfill
\begin{minipage}[t]{0.485\textwidth}
\begin{tcolorbox}[
  colback=analysisbg,
  title=\textbf{Analysis},
  colbacktitle=titlebg,
  coltitle=black,
  fonttitle=\footnotesize,
  height=3.65cm,
  valign=top
]
\footnotesize
#3
\vspace*{\fill}
\end{tcolorbox}
\end{minipage}
}

\usepackage{lineno}

\definecolor{darkblue}{rgb}{0, 0, 0.5}
\hypersetup{colorlinks=true, citecolor=darkblue, linkcolor=darkblue, urlcolor=darkblue}

\title{\textit{Do LLMs Triage Like Clinicians?}
A Dynamic Study of Outpatient Referral}

\author{
\textbf{Xiaoxiao Liu\textsuperscript{1,2}\thanks{Equal contribution.}},
\textbf{Qingying Xiao\textsuperscript{3}\footnotemark[1]},
\textbf{Bingquan Zhang\textsuperscript{1}},
\textbf{Junying Chen\textsuperscript{1}},
\textbf{Xiangyi Feng\textsuperscript{1}}, \\
\textbf{Ziniu Li\textsuperscript{1}},
\textbf{Xiang Wan\textsuperscript{4}},
\textbf{Jian Chang\textsuperscript{2}},
\textbf{Guangjun Yu\textsuperscript{3}},
\textbf{Yan Hu\textsuperscript{1}\thanks{Corresponding author.}},
\textbf{Benyou Wang\textsuperscript{1,4}\footnotemark[2]}
\\
\textsuperscript{1}Chinese University of Hong Kong, Shenzhen, \\
\textsuperscript{2}Bournemouth University, \\
\textsuperscript{3}National Health Data Institute, Shenzhen, \\
\textsuperscript{4}Shenzhen Research Institute of Big Data
\\
\small{\href{mailto:wangbenyou@cuhk.edu.cn}{wangbenyou@cuhk.edu.cn}}
}
\begin{document}

\ifcolmsubmission
\linenumbers
\fi

\maketitle

\begin{abstract}

Outpatient referral (OR) is a core clinical workflow that assigns patients to hospital departments under incomplete and evolving information, yet it is commonly simplified as a static classification problem despite being inherently interactive in practice.
In this work, we study outpatient referral as a dynamic process driven by information acquisition and uncertainty reduction.
We analyze both static scenarios based on fixed patient information and dynamic scenarios involving multi-turn dialogue, to test whether large language models (LLMs) improve referral outcomes through better prediction or more effective questioning.
Our findings show that LLMs offer limited advantages over traditional classifiers in static referral accuracy, but consistently outperform them in dynamic settings by asking discriminative follow-up questions that reduce uncertainty over candidate departments.
These results suggest that the primary value of LLMs in outpatient referral lies not in static prediction, but in supporting interactive, uncertainty-aware clinical decision-making. The code is publicly available at: https://github.com/FreedomIntelligence/IOR-Bench
\end{abstract}

\section{Introduction}

Medical resource allocation is a critical challenge worldwide and a cornerstone of effective healthcare systems~\cite{glasziou2017evidence, gamliel2010attribute}.
Its core objective is to ensure limited medical resources are allocated to the right patients at the right time and place~\cite{yinusa2023optimizing, bravo2021optimization}.
A key component of this process is outpatient referral, which directs patients to appropriate clinical departments, helping reduce overcrowding and improve treatment efficiency.

In real-world practice, outpatient referral decisions are rarely made in a single step.
Instead, clinicians iteratively ask follow-up questions to clarify symptoms and reduce uncertainty before reaching a final decision.
However, most existing studies and evaluation protocols still simplify outpatient referral as a static classification task~\cite{liu2019intelligent}, overlooking its inherently interactive nature.
This mismatch between real-world clinical workflows and existing evaluation paradigms motivates a systematic study of outpatient referral as an independent, interactive decision-making task.
Recent advances have explored the use of LLMs in healthcare~\cite{singhal2023large, nazi2024large}, including outpatient referral~\cite{zhang2023huatuogpt, wan2024outpatient}.
Yet, LLMs struggle to consistently integrate symptoms, medical history, and test results through multi-turn reasoning, leading to incomplete or unstable recommendations in practice.


In this work, we show that outpatient referral is better understood as a dynamic decision-making process driven by uncertainty reduction.
By analyzing referral dialogues through an information-theoretic lens, we find that LLMs offer limited improvements over traditional models in static settings, but distinguish themselves in interactive scenarios by asking follow-up questions that more effectively reduce uncertainty over candidate departments.
These findings suggest that the value of LLMs in outpatient referral lies less in static prediction accuracy and more in their ability to acquire discriminative information through interaction.

The \textbf{contributions} of this paper are as follows.
(i),  we \textbf{reconceptualize} outpatient referral as a dynamic decision-making process driven by information acquisition, rather than a purely static classification task, aligning the problem formulation with real-world clinical practice.
(ii), we provide a systematic empirical study of outpatient referral across both static and interactive settings, revealing a clear distinction between prediction accuracy and information-seeking behavior.
(iii), through an \textbf{information-theoretic analysis} of uncertainty reduction.
(iv), we conduct a comprehensive study  comparing large language models, traditional classifiers, and \textbf{human experts}.

\begin{figure*}[t]
\centering
\vspace{-15pt}
\includegraphics[width=1\linewidth]{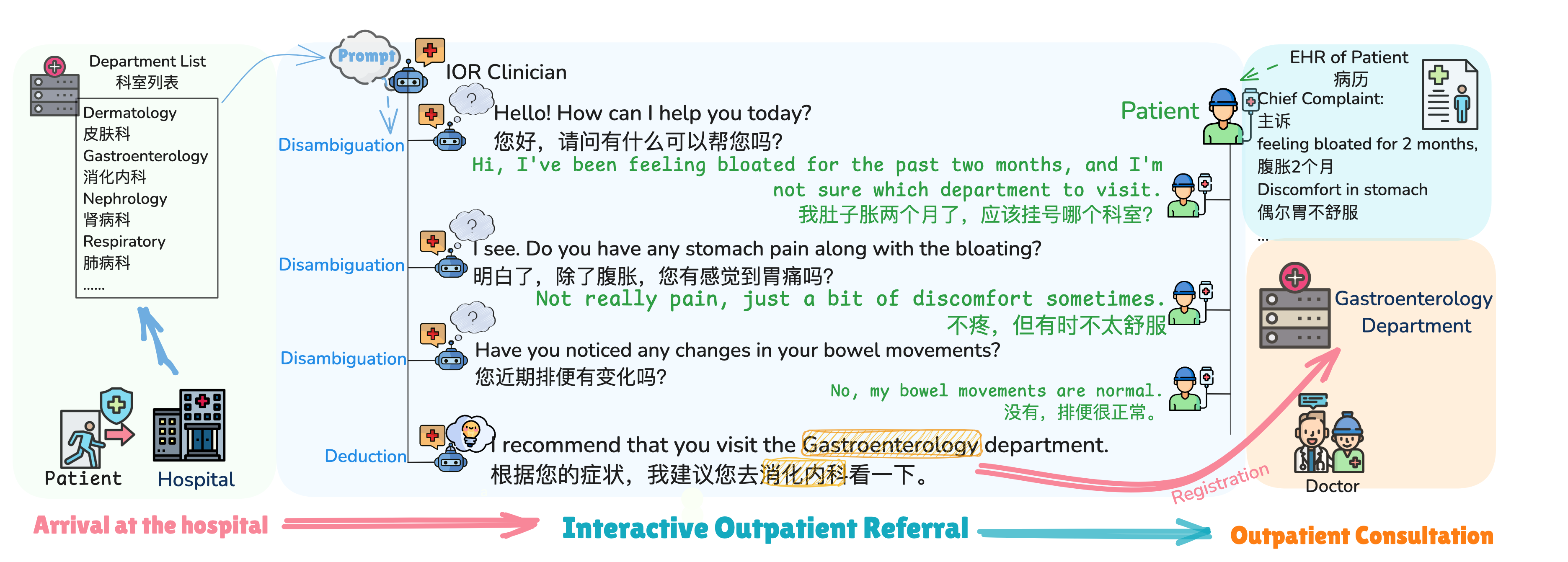}
\vspace{-20pt}
\caption{Pipeline for the process of the outpatient referral task.} 

\label{fig: pipeline}
\end{figure*}

\section{Background}
\label{sec: problemStat}

\subsection{Problem Formulation}

Outpatient referral~\footnote{Healthcare systems differ in how referrals are managed.
In GP-based systems common in Western countries~\cite{safi2022today, freytag2016impact}, General Practitioners act as gatekeepers who conduct preliminary assessments and coordinate specialist referrals~\cite{schmalstieg2021role}.
In contrast, China adopts a \textit{direct-access} system, where patients bypass primary care and directly seek services at large hospitals.
As a result, patients must independently choose specialized departments (e.g., Pulmonology, Nephrology, Endocrinology), a process known as \textit{outpatient referral}} aims to \textit{assign a patient to the most appropriate clinical department under incomplete and evolving information}(Figure~\ref{fig: pipeline}). 
This task is inherently challenging due to overlapping symptoms across departments and the need to make decisions before a full clinical picture is available.
In practice, outpatient referral in direct-access systems has traditionally been performed by nurses at hospital guidance desks, who assess patients based on chief complaints and symptoms.
This process requires broad cross-departmental expertise and is subject to heavy workload, leading to non-negligible error rates~\cite{ge2020design}.
These challenges motivate the use of automated referral systems.

The goal of outpatient referral is to refer a patient to the best-suited department. It is usually framed as  a straightforward \textit{classification} problem:
\begin{component}\label{component:static}
\textbf{Department Classification} aims to classify a patient to a department based on a \textit{static} context (e.g. a single-turn consulting question or a given consulting dialogue).
\end{component}

However, real-world medical consultations are far more complex, often requiring dynamic interactions, especially querying, to clarify symptoms and gather additional information.

\begin{component} \textbf{Proactively  Questioning} involves dynamically asking relevant questions during a dialogue to reduce uncertainty and improve classification accuracy. 
\end{component}
This adaptive approach helps refine the information gathered, leading to more accurate department classification.
This makes generative models powered by LLMs crucial for providing more accurate recommendations; see Table~\ref{tab:questionDef} for a concise comparison of the static and dynamic outpatient referral processes.

\subsection{Outpatient Referral as a Dynamic  Process}

\noindent \textbf{The Naive Static Process}~
Most existing approaches formulate outpatient referral as a static department
classification problem~\cite{singhal2023large, nazi2024large, gaber2025evaluating},
as illustrated in Component~\ref{component:static}. Specifically, outpatient
referral is treated as a one-shot decision process that maps a given dialogue
context to a target department:
\begin{equation}
\small
\label{eq:classification}
f: \mathcal{H} \rightarrow \mathcal{D}.
\end{equation}
Here, $h_0 \in \mathcal{H}$ denotes the initial (historical) dialogue context,
and $\mathcal{D}$ represents the set of departments in a hospital, with
$d \in \mathcal{D}$ being the ground-truth referral decision. This formulation
naturally reduces outpatient referral to a standard classification problem.
However, in practice, the department set $\mathcal{D}$ varies significantly
across hospitals in China, which poses challenges for model generalization and
deployment.

\begin{table}[t]
\centering 

\setlength{\tabcolsep}{1mm}
\small
\begin{tabular}{lll}
\toprule
& \textbf{Static Process} & \textbf{Dynamic Process} \\
\midrule
\textbf{Task} & classification & questioning \\
\textbf{LLM Arch.} & {\tt enc},  {\tt enc-dec}, {\tt dec} & {\tt dec}, {\tt enc-dec} \\
\textbf{Difficulty} & easy & harder \\
\textbf{Output} & predicted  $d$ &refined questions $\{q_1,\cdots, q_n \}$, $d$\\
\textbf{Process} & single-turn inference & iterative refinement, uncertainty reduction \\
\textbf{Evaluation} &  Acc. &  Acc. , Entropy, IG \\
\bottomrule
\end{tabular}
\caption{Comparison between static and dynamic outpatient referral processes. \textit{LLM Arch.}: LLM architecture; \textit{{\tt enc}}: encoder-only model; \textit{{\tt dec}}: decoder-only model; \textit{{\tt enc-dec}}: encoder-decoder model. \textit{Acc.}: accuracy. \textit{IG}: information gain}
\label{tab:questionDef}
\end{table}

\noindent \textbf{Towards a Dynamic Process}~ 
In real-world clinical settings, outpatient referral is rarely a single-step
decision. Instead, it is an iterative process in which clinically relevant
information is progressively revealed through interaction. Before making a
final referral decision (i.e., applying $f$), a system may proactively ask
follow-up questions to elicit missing or ambiguous information from the patient. Formally, in the dynamic setting, the dialogue context is incrementally
expanded as
\begin{equation} 
\small
h^{(n)} = \{h_0, q_1, a_1, \dots, q_n, a_n\},
\end{equation}
where $q_i$ and $a_i$ denote the $i$-th question generated by the model and the
corresponding patient response, respectively. A questioning model $g$ generates
the next question based on the current context, i.e.,
$g(h^{(n)}) = q_{n+1}$, with the objective of acquiring information that is most
useful for referral. The overall goal is to learn an optimal questioning strategy $g^*$ that
maximizes referral accuracy:
\begin{equation}
\small
g^* = \arg\max_{g} \;
\mathbb{E}_{(h_0, d)} \Big[
\mathbb{I}\!\left(
f\!\left(h_0, q_1, a_1, \dots, q_n, a_n\right) = d
\right)
\Big],
\end{equation}
where $\mathbb{I}[\cdot]$ is an indicator function that evaluates whether the
predicted department produced by $f$ matches the ground-truth department $d$.
The classifier $f$ follows the static formulation in
Eq.~\ref{eq:classification}, but operates on a richer, interaction-enhanced
context.

\begin{figure}[b]
{\includegraphics[width=\linewidth]{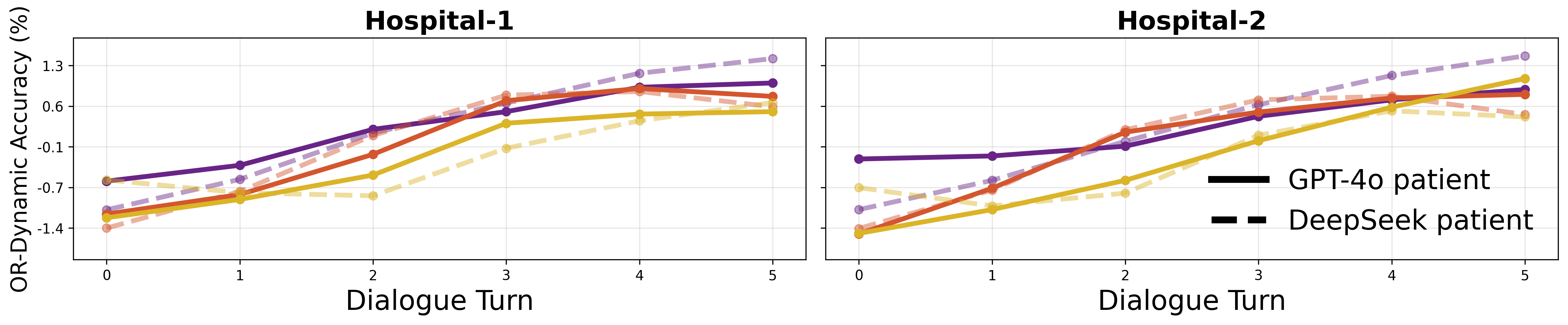}}
\caption{Comparison of dynamic referral accuracy trends under different patient simulator settings.\textit{The curves are smoothed for clarity, while preserving relative distances and trends.}}
\label{fig: vs}
\end{figure}

\noindent \textbf{Interaction with User Simulators}~
To evaluate the dynamic referral process described above, we simulate realistic
outpatient consultations in an interactive setting, as shown in
Figure~\ref{fig: vs}. Two LLM-based agents are engaged in a turn-based dialogue: a
\textbf{Patient Simulator}, which generates responses conditioned on EMR data,
and an \textbf{Outpatient Referral System}, which iteratively asks questions and
predicts the most appropriate department. At each turn, the system generates a
query based on the current context, receives a patient response, and outputs an
interim referral decision, allowing referral accuracy to be tracked as
clinically relevant information is progressively revealed.

\noindent \textbf{Relation between Static and Dynamic Process}~
The generative questioning model $g$ can also be implemented as a
special case of a classification or generation model through appropriate prompt
design. This perspective unifies static classification and interactive
questioning under a single modeling framework.




\subsection{Information Gain in the Dynamic Process}
\label{sec:information_gain}
Real-world outpatient referral decisions are refined through follow-up questioning that elicits ambiguous patient information. Accordingly, we formulate dynamic outpatient referral as a process of reducing uncertainty over candidate departments. Entropy measures the uncertainty of the remaining candidates. Information gain measures how much each interaction reduces it.

\noindent \textbf{Uncertainty over Candidate Departments}~
At dialogue turn $t$, the model maintains a set of candidate departments from the predefined department set $\mathcal{D}$ given the dialogue history $h_t$. The model then assigns each department a relative support score and normalizes it into a distribution $p_t(d)$. Detailed scoring and normalization procedure are provided in Appendix~\ref{app:entropy_details}.

\noindent \textbf{Entropy}~
To quantify decision uncertainty at turn $t$, we compute the Shannon entropy~\cite{shannon1948mathematical} over the resulting department distribution:
\begin{equation} \small
H_t = - \sum_{d \in \mathcal{D}} p_t(d) \log p_t(d).
\end{equation}
Higher entropy indicates greater uncertainty over current information, while lower entropy reflects a more concentrated referral decision. 
\noindent \textbf{Turn-Level Information Gain}~
We define the information gain (IG) at turn $t$ as the reduction in uncertainty:
\begin{equation}  \small
IG_t = H_{t-1} - H_t.
\end{equation}
A positive $IG_t$ indicates that the newly acquired information makes the model's belief over candidate departments more concentrated, thereby reducing decision uncertainty.
If $H_t = H_{t-1}$, then $IG_t = 0$, suggesting that the interaction fails to provide discriminative information for referral.

\section{Benchmark Curation}

\subsection{Case I: Static Evaluation}\label{sec: static-dataset}

\noindent \textbf{Data Overview}~The OR-Static dataset comprises 612 dialogues between real patients and an AI doctor (an LLM deployed in an online outpatient referral system) , within the context of outpatient referral. The OR-Static dataset is sourced from the cooperating institutes and has received approval from the Research Ethics Committees.

\begin{figure*}[b]
\vspace{-20pt}
    \centerline{\includegraphics[width=1\textwidth]{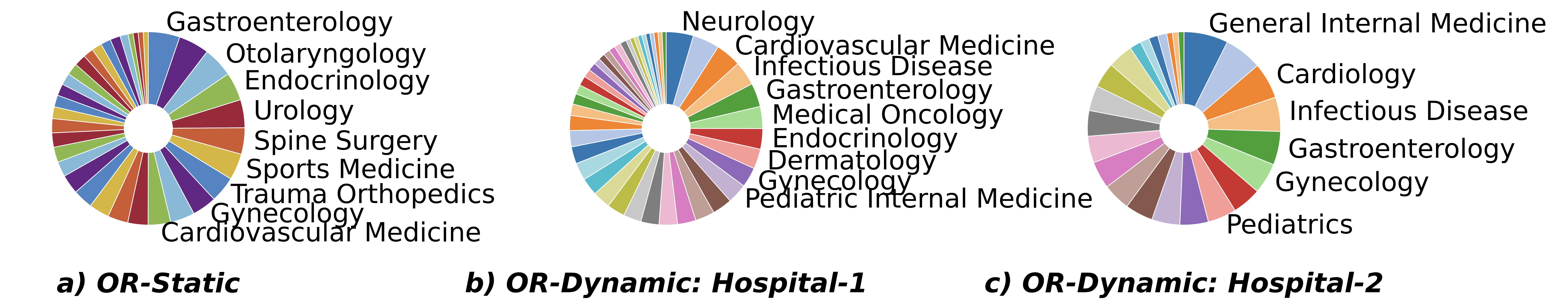}}

    \caption{Department distribution of the OR datasets across the static and dynamic settings. Only the most frequent departments are labeled. For a more detailed department overview, see Appendix~\ref{sec:department_distribution}.
    }
    \label{fig:dept-distribution}
\end{figure*}
\noindent \textbf{Data Curation}~ 
The dataset was first de-identified by hospital staff using rule-based anonymization procedures. It covered outpatient records collected between August and October 2024. The cases were then stratified to ensure balanced representation across departments and demographic groups, refined with GPT-4o to improve linguistic consistency, and finally reviewed by licensed physicians to verify privacy compliance and clinical validity. See Appendix~\ref{apdx:static_curation} for processing details.

\noindent \textbf{Annotation Process}~ 
Two licensed physicians, both with experience in outpatient referral, independently reviewed and annotated the data according to predefined guidelines (Figure~\ref{fig: dynamic-annotation-guid} in Appendix~\ref{apdx:static-guid}), ensuring realistic scenarios, accurate department labels, privacy compliance, and overall coherence. Cases with discrepancies were excluded. Out of 1,000 dialogue scripts, 612 scripts covering 35 departments (Figure~\ref{fig:dept-distribution}-a) met the criteria and were retained. Each sample consists of a real outpatient referral dialogue paired with its corresponding target department.

\noindent \textbf{OR-Static Evaluation Settings}~ 
\label{sec:static-evaluation-setup}~
Given a static dialogue history, the model was asked to match it to a target department. The predicted department is then compared with the annotated ground-truth label.
To evaluate the model's sensitivity in this task to prompt variations, we adopted four distinct prompt strategies (Appendix~\ref{apdx: static-prompt-design}). 
In addition, Macro-F1 scores are reported to better reflect performance across departments under class imbalance (See Table~\ref{tab:staticEvaluation_macroF1} in Appendix~\ref{apdx:staic_macrof1}).


\subsection{Case II: Dynamic Evaluation} \label{sec: Dynamic_eval_data}

\noindent \textbf{Data Overview}~ 
The OR-Dynamic dataset for the dynamic assessment was sourced from the public platform (\url{https://www.yiigle.com/index}).
The publicly available dataset was rigorously anonymized to ensure compliance with privacy and data protection regulations. The dataset includes anonymized electronic medical records (EMRs) of patients. The data structure contains several key fields: \textit{``age'' , ``sex'' , ``chief complaint'' , ``history of present illness'' , ``past medical history'' , ``personal history''  and ``family history'' } along with the department visited by each patient.

\noindent \textbf{Data Curation}~ 
A medical expert first reviewed 100 randomly sampled records to establish the \textit{Relevance}, \textit{Completeness}, and \textit{Quality} criteria (Appendix~\ref{apdx:static_criteria}). Guided by these criteria, GPT-4o extracted 2,000 records that met the defined standards. Two licensed physicians with outpatient referral experience then independently annotated the department labels for these samples using department lists from \textit{Hospital-1} (tertiary care) and \textit{Hospital-2} (secondary care) (Appendix~\ref{apdx:dynamic_curation}).

\noindent \textbf{Data Annotation}\label{sec: dynamic-guidance}~ 
Two licensed physicians with outpatient referral experience annotated the unlabeled data using department lists from two hospitals, following predefined guidelines (See Figure~\ref{fig: dynamic-annotation-guid} in Appendix~\ref{apdx:dynamic-guid}) to ensure consistency and accuracy. Cases with annotation discrepancies were excluded. After validation, 860 high-quality records were retained, covering 45 departments in Hospital-1 and 25 in Hospital-2 (Figure~\ref{fig:dept-distribution}).


\noindent \textbf{OR-Dynamic Evaluation Settings}~ \label{sec:dynamic-evaluation-setup}~
We use LLMs to simulate patients, while the evaluated LLMs act as doctors interacting with them. At the end of each turn, the doctor LLM asks a follow-up question and updates its predicted target department (See Appendix~\ref{apdx: dynamic-prompt-design} for prompt designs). In addition, we conducted an ablation experiments using two alternative LLMs as patient agents for further comparison (Appendix~\ref{apdx:robustness_simulator}).


\subsection{On the Fidelity of Patient Simulators}~
\noindent \textbf{Settings to Compare Human and Simulated Patients}~\label{apdx:sp-simulator}~
To ensure dynamic outpatient referral evaluation reflects the behavior of the~\emph{doctor model}, we invite three physicians to role-play patients and compare their interactions with those generated by multiple LLM-based simulators. Detailed evaluation setup is described in Appendix~\ref{apdx:simulator_fidelity}. We also include an expert fidelity analysis shown in Appendix~\ref{apdx:dilog_N_analss}.

As shown in Figure~\ref{fig:fed_sp}, GPT-4o exhibits the closest alignment with human experts across all four metrics, with comparable dialogue structure and interaction efficiency. Based on physician assessment, GPT-4o is selected as the patient simulator for OR-dynamic evaluation experiments. See Appendix~\ref{apdx:simulator_fidelity} for Detailed scoring criteria and inter-expert agreement.

\begin{figure*}[t]
\vspace{-20pt}
{\includegraphics[width=\linewidth]{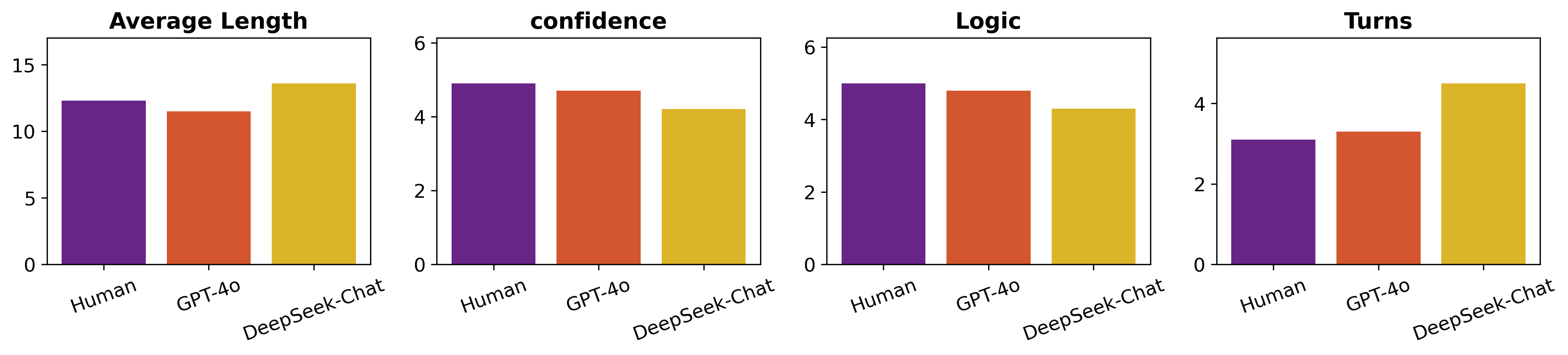}}
\vspace{-15pt}
\caption{Comparison of evaluation settings between human experts simulating patients and LLM agents simulating patients.}
\label{fig:fed_sp}
\end{figure*}

\noindent \textbf{Results}~
\label{apdx:dynamic_compare}

We conducted similar experiments using DeepSeek-Chat as the Patient simulator to evaluate several models in the OR-dynamic task, with the results shown in Table~\ref{tab: compare_dynamic} of Appendix~\ref{apdx:robustness_simulator}. We find that the performance trends of the models remain largely consistent across both tables. This stability indicates that the OR Doctor Agent's performance is robust, regardless of whether GPT-4o or DeepSeek-Chat is used as the Patient Agent, confirming the reliability of the evaluation framework.
\begin{takeaway}
    Overall, the accuracy evolution patterns remain highly consistent across patient simulators, indicating that the observed refinement behavior is not sensitive to the choice of simulator.
\end{takeaway}

\section{Experimental Study on OR Systems} 

\subsection{Experimental Settings}

Most experiments were conducted in October 2024, with several newer models added later. We selected recent, widely used LLMs across different scales and architectures to ensure broad SOTA coverage (see Appendix~\ref{apdx:model-selection} for the selection criteria). The main results are presented below, while additional analyses of language characteristics, prompt strategies, and hospital-specific trends are provided in Appendix~\ref{sec:language-analysis}, Appendix~\ref{apdx:prompt-analysis}, and Appendix~\ref{apdx:hospital-analysis}.

\subsection{Results and Analysis}

\begin{figure}[h]
\vspace{-15pt}
\centering

\includegraphics[width=\linewidth]{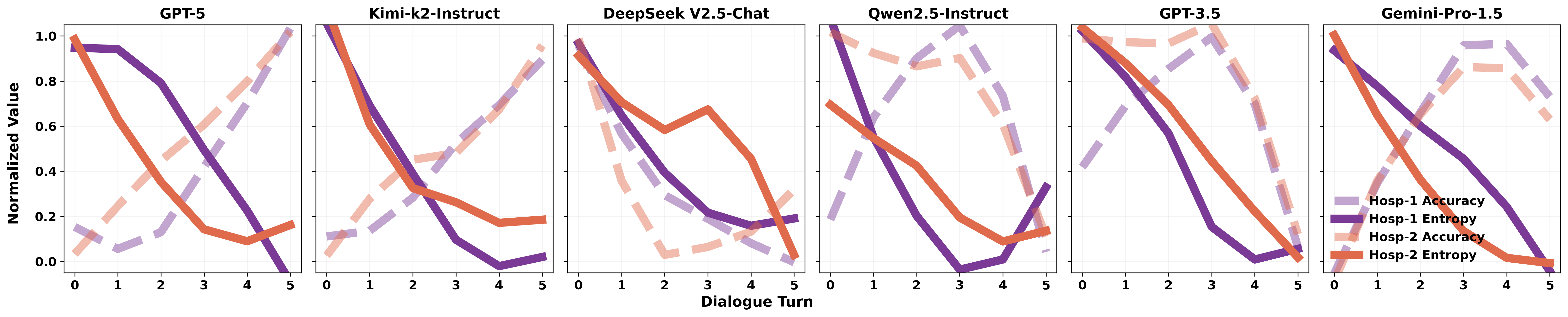}
\caption{Accuracy (solid line) and entropy (dashed line) curves over dialogue turns for selected models. Values are normalized and smoothed for trend comparison. For the comparison of all models, see Figure~\ref{fig:ig_all} in Appendix~\ref{apdx:entropy}} 

\label{fig: infogain}
\end{figure}

\noindent \textbf{Entropy}~
From Figure~\ref{fig: infogain}, uncertainty reduction does not consistently translate into accuracy improvement, as the sharp entropy declines shown in GPT-3.5 are accompanied by weak curves at later turns. In contrast, stronger models such as GPT-5 and often improve accuracy with smaller entropy changes, suggesting that their follow-up questions are discriminative rather than aggressively space-reducing. Effective referral therefore depends more on asking discriminative questions than on reducing uncertainty (See Appendix~\ref{apdx:entropy} for more details).

\begin{takeaway}
    Effective OR depends more on asking discriminative questions than on aggressively reducing uncertainty.
\end{takeaway}

\noindent \textbf{Static Results vs. Dynamic Results}~ 
In static tasks, LLMs lack competitiveness against traditional models, as BERT performs comparably to or even outperforms many LLMs (Table~\ref{tab: staticEvaluation}). However, BERT cannot participate in dynamic evaluation, which requires multi-turn interactions. In these settings, LLMs show steady improvement over turns, suggesting that interactive questioning enables better decision refinement (Table~\ref{tab: dynamicResult}).

\begin{takeaway}
LLMs provide limited advantages over non-LLM models in static classification tasks but become beneficial in dynamic settings through interactive questioning.

\end{takeaway}

\begin{table*}[t]
\centering
\small
\setlength{\tabcolsep}{10pt}
\resizebox{\textwidth}{!}{
\begin{tabular}{l|r|rrrr|c}
\toprule
\textbf{Model} & \small\textbf{\#Params (B)} 
& \textbf{Zero-shot$\uparrow$} 
& \textbf{Few-shot$\uparrow$} 
& \textbf{Majority Vote$\uparrow$} 
& \textbf{CoT$\uparrow$} 
& \textbf{Avg.} \\ 
\midrule

\rowcolor{gray!10} \multicolumn{7}{c}{\textit{Medical LLMs}} \\ 
\midrule

\textbf{BianQue-2} & 6.5B
& $17.81 \pm 5.62$ & $8.17 \pm 4.93$ & $16.34 \pm 5.21$ & $4.74 \pm 3.12$ & 11.77 \\

\textbf{DISC-MedLLM} & 13B
& $50.33 \pm 5.89$ & $43.46 \pm 6.32$ & $59.80 \pm 6.01$ & $56.05 \pm 6.46$ & 52.41 \\

\textbf{HuatuoGPT-II} & 34B
& $\mathbf{74.55 \pm 1.92}$ & $\mathbf{73.90 \pm 1.74}$ & $\mathbf{74.47 \pm 1.68}$ & $\mathbf{70.30 \pm 1.66}$ & \textbf{73.30} \\

\textbf{Apollo} & 72B
& $66.18 \pm 5.42$ & $54.58 \pm 5.99$ & $70.10 \pm 5.61$ & $64.05 \pm 6.81$ & 63.73 \\

\midrule
\rowcolor{gray!10} \multicolumn{7}{c}{\textit{English-centric LLMs}} \\ 
\midrule

\textbf{Gemma-Instruct} & 9B
& $67.48 \pm 1.31$ & $67.97 \pm 1.22$ & $70.59 \pm 1.18$ & $69.78 \pm 1.36$ & 68.96 \\

\textbf{Mixtral-8x-Instruct} & 22B
& $64.12 \pm 2.94$ & $65.19 \pm 2.81$ & $68.72 \pm 2.76$ & $60.66 \pm 2.98$ & 64.67 \\

\textbf{Llama-3.2-Instruct} & 90B
& $65.85 \pm 3.34$ & $65.20 \pm 3.18$ & $72.55 \pm 3.07$ & $64.48 \pm 3.33$ & 67.02 \\

\textbf{GPT-3.5} & 175B
& $66.23 \pm 1.02$ & $68.67 \pm 0.94$ & $66.56 \pm 0.91$ & $67.69 \pm 0.97$ & 67.29 \\

\textbf{GPT-4o} & --
& $73.44 \pm 1.19$ & $75.16 \pm 1.35$ & $75.29 \pm 1.16$ & $72.29 \pm 1.29$ & 74.05 \\

\textbf{o1-mini} & --
& $76.14 \pm 1.13$ & $75.33 \pm 1.09$ & $76.80 \pm 1.06$ & $76.47 \pm 1.12$ & 76.19 \\

\textbf{o1} & --
& $\mathbf{78.76 \pm 0.47}$ & $77.45 \pm 0.61$ 
& $77.45 \pm 0.59$ & $77.95 \pm 0.49$ & 77.90 \\

\textbf{GPT-5} & -- 
& $\mathbf{78.76 \pm 0.47}$ & $\mathbf{79.90 \pm 0.50}$ & $\mathbf{79.91 \pm 0.48}$ & $\mathbf{80.23 \pm 0.45}$ & \textbf{79.70} \\

\textbf{Claude-3.5-Sonnet} & --
& $77.47 \pm 4.62$ & $70.84 \pm 4.38$ & $77.21 \pm 4.41$ & $66.67 \pm 4.75$ & 73.05 \\

\textbf{Gemini-Pro-1.5} & --
& $73.05 \pm 2.71$ & $72.08 \pm 2.63$ & $69.57 \pm 2.54$ & $66.23 \pm 2.68$ & 70.23 \\

\midrule
\rowcolor{gray!10} \multicolumn{7}{c}{\textit{Chinese-centric LLMs}} \\ 
\midrule

\textbf{Kimi-k2-Instruct} & 32B 
& $\mathbf{81.69 \pm 0.44}$ & $\mathbf{81.70 \pm 0.46}$ & $\mathbf{82.03 \pm 0.43}$ & $76.31 \pm 0.51$ & \textbf{80.43} \\

\textbf{Qwen3} & 32B 
& $76.80 \pm 0.48$ & $75.16 \pm 0.52$ & $76.96 \pm 0.50$ & $73.53 \pm 0.55$ & 75.61 \\

\textbf{Yi-1.5-Chat} & 34B
& $66.72 \pm 0.45$ & $43.99 \pm 0.39$ & $70.75 \pm 0.44$ & $55.75 \pm 0.36$ & 69.30 \\

\textbf{Qwen2.5-Instruct} & 72B
& $78.10 \pm 0.39$ & $77.94 \pm 0.46$ & $77.78 \pm 0.51$ & $78.92 \pm 0.40$ & 78.19 \\

\textbf{DeepSeek-2.5-Chat} & 236B
& $76.50 \pm 0.98$ & $74.65 \pm 0.93$ & $76.18 \pm 0.89$ & $74.23 \pm 0.97$ & 75.39 \\

\textbf{DeepSeek-r1} & 617B
& $79.58 \pm 0.44$ & $79.41 \pm 0.42$ & $79.97 \pm 0.41$ & $78.76 \pm 0.46$ & 79.43 \\

\textbf{DeepSeek-v3} & 617B
& $79.58 \pm 0.44$ & $80.39 \pm 0.52$ 
& $80.23 \pm 0.51$ & $\mathbf{80.88 \pm 0.41}$ & 80.27 \\

\textbf{Moonshot-v1-128k} & --
& $79.41 \pm 1.78$ & $76.14 \pm 1.69$ & $80.72 \pm 1.61$ & $77.45 \pm 1.76$ & 78.43 \\

\midrule
\multicolumn{7}{c}{\textit{Traditional Classification Models}} \\ 
\midrule
\textbf{Bert-Base-Chinese} & -- & \multicolumn{5}{c}{\textbf{76.00}} \\

\midrule
\multicolumn{7}{c}{\textit{Human Baselines}} \\ 
\midrule
\textbf{Doctors} & -- & \multicolumn{5}{c}{87.24} \\
\textbf{Nurses} & -- & \multicolumn{5}{c}{85.11} \\
\textbf{Laypersons} & -- & \multicolumn{5}{c}{55.22} \\

\bottomrule
\end{tabular}
}
\caption{\textbf{Static evaluation} Top-1 accuracy (\%) across different prompting strategies. ``$\pm$'' denotes 95\% confidence intervals over five independent runs.
Avg.\ denotes the mean accuracy across all prompting strategies. 
``--'' indicates models with undisclosed parameter counts. }
\label{tab: staticEvaluation}
\end{table*}

\noindent \textbf{Scalability vs. Efficiency}~\label{par:scalability} 

From Table~\ref{tab:fewshot}, Bert-Base-Chinese~\cite{fu2024monarch} improves with scaling but remains inefficient, Qwen2.5-3B is more efficient but saturates early, and GPT-4o performs strongly in context without fine-tuning yet quickly plateaus, suggesting limited scalability despite its large pretrained capacity (see Appendix~\ref{apdx:bert_finetuning} for experiment settings).
\begin{takeaway}
    Scalability is not inherent to model size, but to architecture and learning paradigm.
\end{takeaway}

\begin{table*}[h]
\centering
\small
\setlength{\tabcolsep}{1mm}  
\begin{tabular}{llcccccccc}
\toprule
\textbf{Model} & \textbf{Method/Setting} & \multicolumn{8}{c}{\textbf{Performance Across Increasing Data Examples}} \\
\cmidrule{3-10}
& & 50 & 100 & 150 & 200 & 250 & 300 & 350 & 400 \\
\midrule
\textbf{Bert-Base-Chinese} & Supervised Fine-Tuning & 8.06 & 22.58 & 24.19 & 50.81 & 59.68 & 68.55 & 71.77 & \textbf{75.81} \\
\midrule

\textbf{Qwen2.5-3B} & Supervised Fine-Tuning & 42.74 & 60.48 & 64.52 & 67.74 & 75.81 & \textbf{77.42} & 76.61 & 75.81 \\
\midrule

\textbf{GPT-4o} & Few-Shot Learning & 75.61 & 76.42 & 77.05 & 77.23 & 77.87 & \textbf{78.69} & 76.23 & 75.41 \\
\bottomrule
\end{tabular}
\caption{Performance of different models under different learning paradigms at increasing supervision scales (50–400).}
\label{tab:fewshot}
\end{table*}

\subsection{Challenges}~\label{sec:challenges} 
\begin{figure}[h]
\vspace{-10pt}
\centerline{\includegraphics[width=\columnwidth]{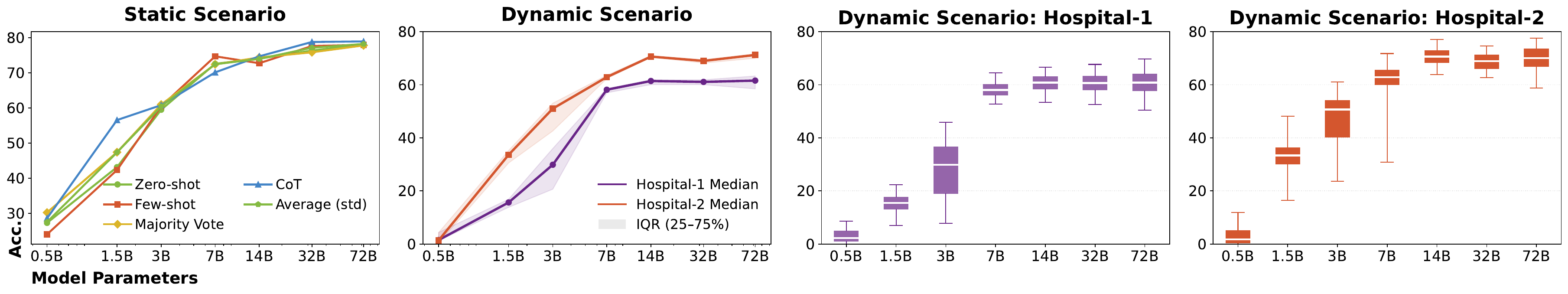}}
\caption{Model performance across different scales (parameter sizes) in static and dynamic scenarios.
The left pair shows static vs. dynamic trends across parameter scales, while the right pair summarizes the dynamic setting with per-parameter boxplots for Hospital‑1 and Hospital‑2; in the dynamic trend plot the line is the median and the shaded band is the interquartile range.}
\label{fig:scaling_law}
\end{figure}
\noindent \textbf{Saturation in Scaling Law}~ 
From Figure~\ref{fig:scaling_law}, \textbf{accuracy} improves as model size increases, but the gains saturate beyond certain threshold. Notably, smaller models are more sensitive to prompt variations, showing larger variation across prompts. Larger models also exhibit better \textbf{stability}, as indicated by the narrower interquartile ranges in the dynamic boxplots. But stability no longer improves and instead declines after a certain scale. See Appendix~\ref{apdx:scaling-law-details} for experimental settings and detailed analysis.

\begin{takeaway}
Scaling LLM sizes improves performance, but gains plateau after a certain threshold.
\end{takeaway}

\begin{table*}[h]
\small
\vspace{-10pt}
\setlength{\tabcolsep}{4pt}
\resizebox{\textwidth}{!}{
\begin{tabular}{l|r|rrrrrr|rrrrrr}
\toprule \multicolumn{1}{l|}{\textbf{Models}}  & \multicolumn{1}{c|}{\textbf{\#Params (B)}}& \multicolumn{6}{c|}{\textbf{Hospital-1}~(45 Departments)}                                    & \multicolumn{6}{c}{\textbf{Hospital-2}~(25 Departments)}                                     \\ \midrule
\textbf{Turn}     &                  & \textbf{0} & \textbf{1} & \textbf{2} & \textbf{3} & \textbf{4} & \textbf{5} & \textbf{0} & \textbf{1} & \textbf{2} & \textbf{3} & \textbf{4} & \textbf{5} \\ \midrule
\rowcolor{gray!10} \multicolumn{14}{c}{\textit{Medical LLMs}}\\ \midrule
\textbf{BianQue-2} & 6.5B & 0.12 & 0.12 & 0.23 & 0 & 0 & 0 & 0.69 & 0.12 & 0.46 & 0.23 & 0.23 & 0 \\
\textbf{DISC-MedLLM} & 13B & 3.09 & 10.65 & 7.06 & 6.25 & 5.90 & 6.71 & 41.32 & 8.56 & 6.83 & 3.94 & 4.05 & 4.51 \\
\textbf{HuatuoGPT-II} & 34B & 43.29 & 32.75 & \textbf{63.31} & 64.47 & 62.62 & 62.38 & 28.82 & 38.19 & 69.21 & \textbf{71.30} & 67.48 & 66.56 \\
\textbf{Apollo} & 72B & 48.15 & 38.08 & 34.61 & 32.29 & 28.24 & 26.50 & 57.13 & 41.02 & 37.43 & 34.99 & 34.88 & 28.97 \\

\midrule
\rowcolor{gray!10} \multicolumn{14}{c}{\textit{English-centric LLMs}}\\ \midrule

\textbf{Gemma-Instruct} & 9B & 50.24 & 50.59 & 52.13 & 51.54 & 47.27 & 41.00 & 62.20 & 64.05 & 65.20 & 65.32 & 59.65 & 52.49 \\
\textbf{Mixtral-Instruct} & 22B & 44.99 & 43.40 & 41.42 & 35.22 & 28.50 & 21.50 & 65.09 & 61.39 & 58.27 & 54.22 & 45.43 & 30.52 \\
\textbf{Llama-3.2-Instruct} & 90B & 46.06 & 46.33 & 46.72 & 42.26 & 44.88 & 47.64 & 61.50 & 55.95 & 59.19 & 56.76 & 56.53 & 58.50 \\
\textbf{GPT-3.5} & 175B & 49.71 & 50.98 & 52.14 & 53.29 & 52.14 & 46.82 & 54.80 & 54.91 & 54.45 & 54.91 & 54.80 & 48.79 \\
\textbf{GPT-4o} & - & 62.43 & 64.05 & 65.20 & 66.59 & 66.70 & 67.89 & 72.83 & 73.99 & 74.91 & 75.61 & 76.00 & 76.99 \\

\textbf{o1-mini} & - & 59.72 & 63.78 & 65.86 & 66.55 & 68.06 & 68.06 & 72.22 & 72.69 & 73.26 & 72.80 & 73.84 & 75.35  \\

\textbf{o1} & - & 64.07 & 64.65 & 66.74 & 66.63 & 68.60 & 70.00 & 71.73 & 75.47 & 74.07 & 75.70 & 74.77 & 75.12 \\

\textbf{GPT-5} & - & 65.80 & 64.32 & 66.10 & 67.24 & 71.34 &\textbf{73.10} & 71.68 & 74.66 & 76.51 & 76.96 & 80.23 & \textbf{81.64}  \\

\textbf{Claude-3.5-Sonnet} & - & 44.16 & 36.53 & 42.20 & 43.70 & 43.70 & 46.40 & 56.18 & 47.51 & 52.49 & 58.73 & 59.19 & 57.46 \\
\textbf{Gemini-Pro-1.5} & - & 55.67 & 56.48 & 57.87 & 59.03 & 58.91 & 58.10 & 65.89 & 66.59 & 68.21 & 69.59 & 68.44 & 68.44 \\

\midrule
\rowcolor{gray!10} \multicolumn{14}{c}{\textit{Chinese-centric LLMs}}
\\ \midrule

\textbf{Kimi-k2-Instruct} & 32B & 64.93 & 66.56 & 65.43 & 66.87 & 68.91 & 67.90 & 74.14 & 75.26 & 74.62 & 76.18 & 75.46 & 76.81 \\

\textbf{Qwen3} & 32B & 67.16 & 67.28 & 67.40 & 64.07 & 68.10 & 67.24
& 75.35 & 74.98 & 76.51 & \textbf{77.00} & 75.21 & 76.44  \\

\textbf{Yi-1.5-Chat} & 34B & 51.45 & 8.32 & 6.13 & 5.20 & 5.09 & 5.43 & 64.05 & 16.76 & 20.46 & 13.64 & 14.34 & 13.87 \\

\textbf{Qwen2.5-Instruct} & 72B & 57.83 & 60.51 & 63.53 & 64.00 & 62.68 & 56.10 & 71.94 & 71.32 & 71.81 & 69.61 & 71.20 & 64.46 \\

\textbf{DeepSeek V2.5-Chat} & 236B & 62.08 & 50.97 & 46.86 & 45.41 & 41.91 & 40.58 & 64.78 & 54.78 & 49.57 & 50.43 & 53.48 & 53.48 \\

\textbf{DeepSeek-r1} & 671B &  58.45 & 57.99 & 60.19 & 59.49 & 59.61 & 60.88 & 59.95 & 59.72 & 60.88 & 60.42 & 58.91 & 60.19 \\

\textbf{DeepSeek-v3} & 671B & 59.26 & 59.49 & 59.82 & 60.76 & 57.06 & 55.09 &69.33 & 67.94 & 68.40 & 66.20 & 63.08 & 57.52 \\                              

\textbf{Moonshot-v1-128k} & - & 
56.94 & 56.60 & 59.72 & 61.69 & \textbf{64.24} & 64.12 & 68.76 & 67.01 & 69.91 & 67.25 & 67.94 & 69.56
\\
 \midrule
\multicolumn{14}{c}{\textit{Human Baselines}} \\ \midrule
\textbf{Doctors} & 
\multicolumn{11}{c}{82.05}
 \\
 \textbf{Nurses} & 
\multicolumn{11}{c}{69.10}
 \\
 \textbf{Laypersons} & 
\multicolumn{11}{c}{37.10}
 \\

 \bottomrule
\end{tabular}
}
\caption{\textbf{Dynamic Evaluation} performance (accuracy\%), with Hospital-1 and Hospital-2 representing tasks of different department lists. Columns 0–5 means dialogue turns.
}
\label{tab: dynamicResult}
\end{table*}

\subsection{Human Experts vs. LLMs}

\noindent \textbf{Experimental Settings}~\label
To compare humans and LLMs under identical conditions, 6 human evaluators completed both static referral on 31 sampled dialogue histories (Appendix~\ref{apdx: static-human-eval}) and dynamic referral on 30 sampled cases via interaction with LLM-simulated patients (Appendix~\ref{apdx: dynamic-human-eval}).

\begin{figure}[h]
\centering
\vspace{-10pt}
\centerline{\includegraphics[width=0.9\columnwidth]{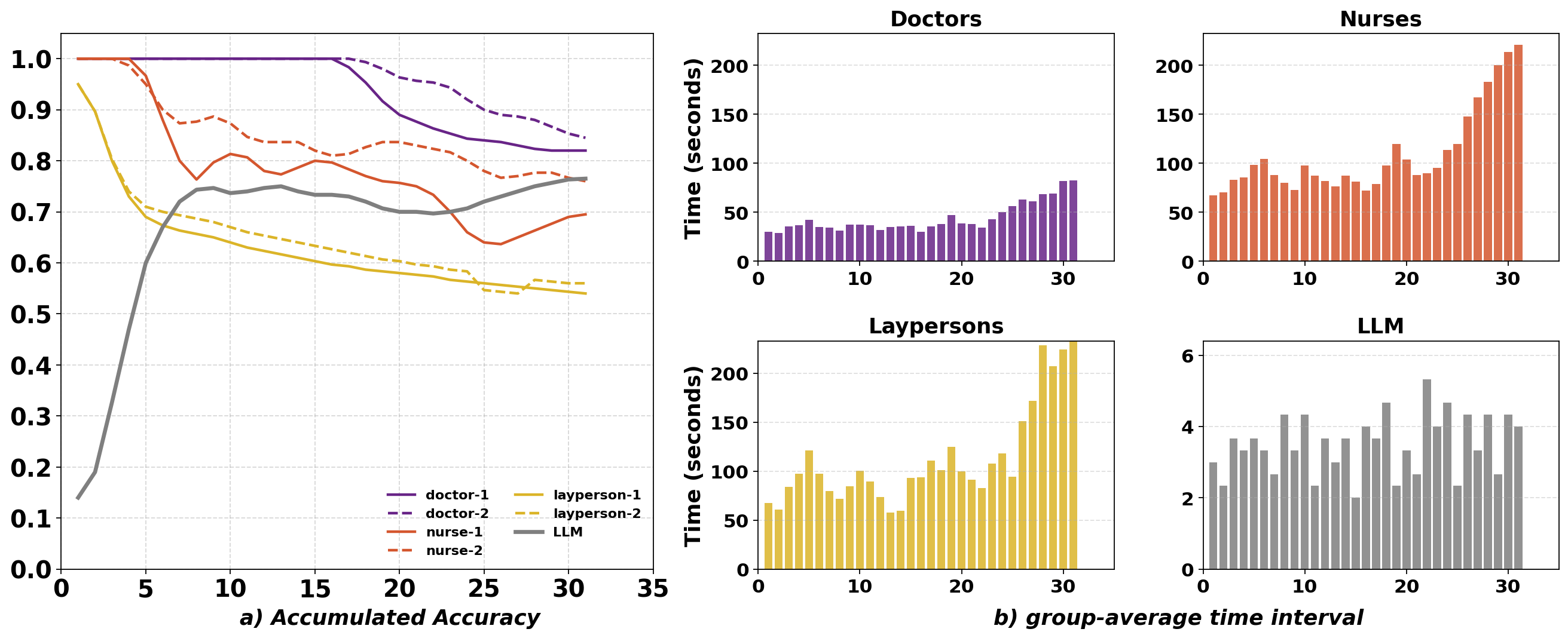}}
\caption{Comparison of accumulated accuracy and group-average time-interval dynamics across groups, with time intervals smoothed for readability.}
\label{fig:cumulative_accuracy}
\end{figure}

\noindent \textbf{Accuracy Comparison}~
As shown in Figure~\ref{fig:cumulative_accuracy}-a, doctors achieved the highest accuracy in the static task, while nurses performed comparably to LLMs. In the dynamic task, as illustrated in Table~\ref{tab: dynamicResult}, humans excelled in scenarios requiring contextual reasoning, but their performance varied significantly depending on their professional background. 

\begin{takeaway}

SOTA LLMs outperform laypersons, but far behind professional nurses and doctors.
\end{takeaway}

\noindent \textbf{Behavioral Difference}~
We analyze how referral accuracy evolves over consultation time when humans and LLMs act as doctors in outpatient referral tasks (see Appendix~\ref{apdx: dynamic-human-eval} for more setting details).
Human experts exhibit a clear performance degradation over time: for instance, \texttt{Doctor-1} experiences an accuracy drop of approximately 50\%, with nurses and laypersons showing even larger declines, as shown in Figure~\ref{fig:cumulative_accuracy}-a which shows the accuracy over cumulative cases (See Appendix~\ref{apdx:acc-clarification} and Appendix~\ref{apdx:acc-vis} for more details).
In contrast, LLMs maintain stable Figure~\ref{fig:cumulative_accuracy}-b, time-invariant performance across consultations, highlighting a key behavioral difference between human and model-based referral systems and suggesting a practical advantage of LLMs in repetitive workloads~\cite{xiao2024densing}.
\begin{takeaway}

While human experts initially outperform LLMs, their performance degrades over time probably due to fatigue, whereas LLMs maintain stable and consistent performance, making them well suited for repetitive tasks.

\end{takeaway}


\section{Conclusion}
This study examines large language models in outpatient referral under both static and interactive settings. We find that LLMs offer limited benefit in static classification, but become more effective in dynamic referral by asking discriminative follow-up questions. More broadly, healthcare AI should be evaluated not only by final accuracy, but also by how models gather and use information through interaction.

\newpage

\section*{Limitation}~\label{apdx:limitation-future}

\noindent \textbf{On the language}~
This study is based on Chinese language data and focuses on the Outpatient Referral system in Chinese hospitals. Unlike the Western healthcare system, which relies heavily on family physicians, the Outpatient Referral system in Chinese hospitals emphasizes structured dialogue interactions and department-based OR. However, we believe that the core problem definition proposed in this study has broad applicability. This dialogue-based information acquisition and classification model is not only suitable for Outpatient Referral but can also be extended to other consultation-based problem-solving contexts. We hope that the findings of this research will not only provide practical guidance for consultation systems in Chinese contexts but also offer valuable insights and inspiration for consultation scenarios in other languages.

\noindent  \textbf{Experimental Settings}~
In terms of static dataset annotation, the evaluation of dialogue coherence, including fluency, logical flow, and emotional expression, was not fully implemented. This task remains an area for improvement in future work.
While OR-Bench provides a comprehensive framework for evaluating outpatient referral tasks, it still needs to be discussed in conjunction with other scenarios, such as handling rare or ambiguous medical cases. Additionally, the model’s ability to generate appropriate follow-up questions in dynamic dialogues remains inconsistent, showing varying performance that requires further improvement.

\section*{Ethics Statement}

The study was approved by the institute on August 30, 2025. All annotators were fairly compensated, adhering to the standard hourly wage practices of their respective states.

\bibliography{colm2026_conference}
\bibliographystyle{colm2026_conference}

\appendix

\section{Related Work}


\subsection{Outpatient Referral}
Previous studies using medical knowledge graphs for predicting patient department visits from electronic healthcare records (EHRs) \cite{liu2019intelligent, yu2021medical} suffer from low accuracy and reliance on manual annotation. LLMs have improved intelligent healthcare systems \cite{thirunavukarasu2023large, lambert2023integrative}, especially in optimizing outpatient referrals \cite{zhang2023trinity, wan2024outpatient, taylor2024bespoke}. These systems analyze patient-reported symptoms to accurately direct patients to the right specialist \cite{mu2024potential, chen2024huatuogpt, lievin2024can, nerella2024transformers}. Key challenges include ensuring precise recommendations and asking clarifying questions to handle vague or incomplete patient inputs~\citep{clusmann2023future, kim2023chatgpt, cascella2023evaluating}.

\subsection{Medical Benchmark}

Recent research on Chinese medical LLMs has focused on QA tasks. \cite{liu2024medbench} introduced ``MedBench,'' a benchmarking system with a large dataset and cloud-based infrastructure, ensuring unbiased, reproducible results aligned with medical professionals. However, it emphasizes single-turn, static QA and lacks support for multi-turn interactions. \cite{wang-etal-2024-cmb} addressed this with the Comprehensive Medical Benchmark (CMB), expanding coverage of clinical QA in Chinese. Still, it remains limited to non-interactive settings and does not assess reasoning across dialogue turns.

\cite{Chen2024CoDTA} proposed generating synthetic patient cases from disease encyclopedias to reduce annotation costs. While effective for pretraining, the generated content lacks the ambiguity and variability of real-world interactions and is not designed for downstream tasks like dynamic triage. Overall, these works highlight a growing interest in Chinese medical LLM evaluation, but none simulate the interactive, multi-turn reasoning essential for clinical decision-making—underscoring the need for frameworks like IOR-Bench.

\section{Entropy Calculation}
\label{app:entropy_details}

Given the dialogue history $h_t$, the model assigns a support score $s_t(d)$ to each department $d \in \mathcal{D}$ according to a fixed clinical scoring rubric written by physicians. Higher scores indicate that the currently available information provides stronger support for that department.

These support scores are converted into a normalized distribution using a softmax transformation:
\begin{equation} \small
p_t(d)=\frac{\exp \left(s_t(d)/\tau \right)}{\sum_{d' \in \mathcal{D}} \exp \left(s_t(d')/\tau \right)},
\end{equation}
where $\tau$ is a temperature parameter.

To reduce the influence of the low-probability tail, departments are ranked by $p_t(d)$ in descending order, and the smallest candidate set $\mathcal{C}_t \subseteq \mathcal{D}$ is retained such that
\begin{equation} \small
\sum_{d \in \mathcal{C}_t} p_t(d) \ge \alpha.
\end{equation}

Entropy is then computed after renormalizing the retained probability mass:
\begin{equation} \small
\tilde{p}_t(d)=\frac{p_t(d)}{\sum_{d' \in \mathcal{C}_t} p_t(d')}, \quad d \in \mathcal{C}_t,
\end{equation}
\begin{equation} \small
H_t = - \sum_{d \in \mathcal{C}_t} \tilde{p}_t(d) \log \tilde{p}_t(d).
\end{equation}

\section{OR-Static Dataset Guidance}\label{apdx:static-guid}
\begin{figure*}[htbp]
\centerline{\includegraphics[width=1\textwidth]{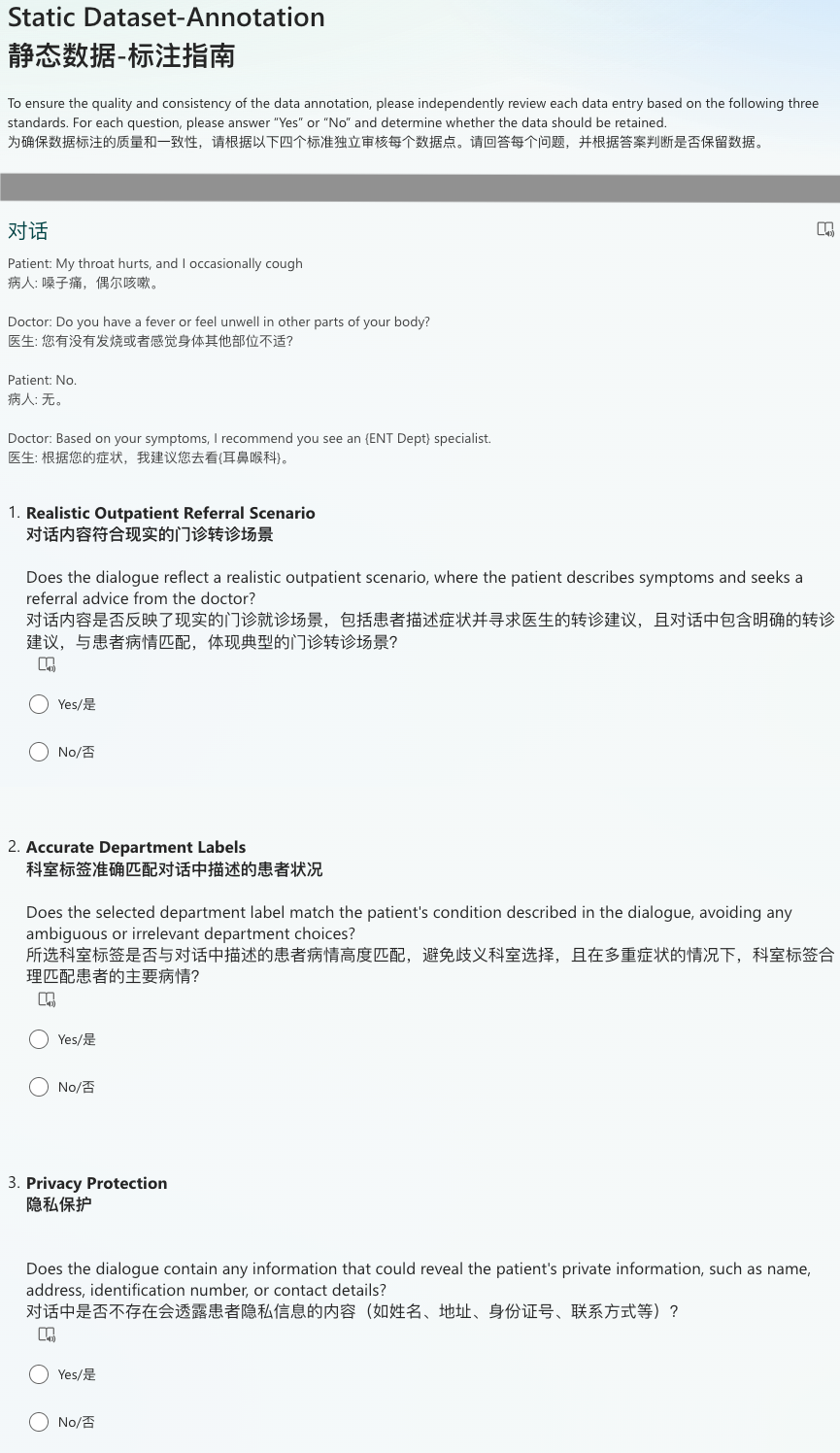}}
\caption{Static dataset annotation guidance 
} 
\label{fig: static-annotation-guid}
\end{figure*}

The guidance shown in Figure~\ref{fig: static-annotation-guid} outlines a framework for annotating dynamic medical datasets, focusing on the completeness, accuracy, and coherence of patient records. It evaluates critical sections, including main symptoms, history of present illness, and past medical history, ensuring these are detailed, specific, and concise. The emphasis is on avoiding vague or fragmented expressions while providing comprehensive documentation. Each section is reviewed through clear, binary criteria to assess adherence to these standards.

\section{OR-Static Data Curation Process}
\label{apdx:static_curation}

\noindent \textbf{Rule-based Anonymization.}
All identifiable information (names, phone numbers, ID numbers, addresses, and temporal markers) was removed using regular-expression–based scripts developed by the hospital’s IT department. The anonymization procedure followed institutional privacy policies and was verified under ethics approval. Each record was manually spot-checked to ensure no residual identifiers remained.

\noindent \textbf{Stratified Sampling.}
Cases were stratified across 12 clinical departments and key demographic attributes (sex, age group) to approximate real-world referral distributions. Sampling ratios were determined from hospital-level outpatient statistics to mitigate overrepresentation of common departments such as Internal Medicine.

\noindent \textbf{GPT-4o Refinement.}
To improve linguistic uniformity without altering clinical meaning, each dialogue was refined with GPT-4o using a structured prompt that constrained editing to stylistic normalization. The prompt explicitly instructed the model to preserve all factual and medical details. A random 10\% subset was manually inspected to confirm adherence.

\noindent \textbf{Physician Review.}
Two licensed physicians from General Internal Medicine and Family Medicine, both with more than five years of outpatient referral experience, independently reviewed the refined scripts. They verified privacy compliance, diagnostic plausibility, and inter-departmental labeling accuracy. Discrepancies were resolved by discussion, and cases failing consensus were excluded.

\section{OR-Dynamic Data Curation Criteria}\label{apdx:static_criteria}
A medical expert reviewed an initial sample of 100 records and established the following selection criteria:
\begin{itemize}
\item \textbf{Relevance}: Records must include clear and specific chief complaints.
\item \textbf{Completeness}: Records must include complete and detailed information on history of present illness, past medical history, and family history.
\item \textbf{Quality}: Include only well-structured and clearly documented records.
\item \textbf{Consistency}: Cases with annotation disagreement were excluded, retaining only samples where two independent medical experts fully agreed on labels. This process minimized ambiguity and ensured high data reliability.
\end{itemize}

For annotation standardization, two board-certified medical experts independently labeled all cases. Divergent annotations underwent automatic exclusion, with only concordant cases retained in the final dataset. This rigorous approach guaranteed unambiguous, clinically consistent data for downstream analysis.

\section{OR-Dynamic Dataset Guidance}\label{apdx:dynamic-guid}
The guidance shown in Figure~\ref{fig: dynamic-annotation-guid} focuses on annotating and evaluating EMRs for the OR-Dynamic datasets. It emphasizes reviewing three key aspects of medical records: the accuracy and detail of the main symptoms, the completeness of the history of present illness, and the coherence of the past medical history. These evaluations are aimed at ensuring precise and comprehensive documentation without fragmented or vague expressions. It also introduces scenarios requiring the selection of appropriate medical departments based on the patient’s symptoms and condition. Two hospital-specific department lists are provided, with evaluators tasked with matching the patient's condition to the most suitable department. This guidance combines structured assessments of medical records with dynamic decision-making for department recommendations, promoting clarity and accuracy in medical documentation and referral processes.

\begin{figure*}[htbp]
\centerline{\includegraphics[width=1\textwidth]{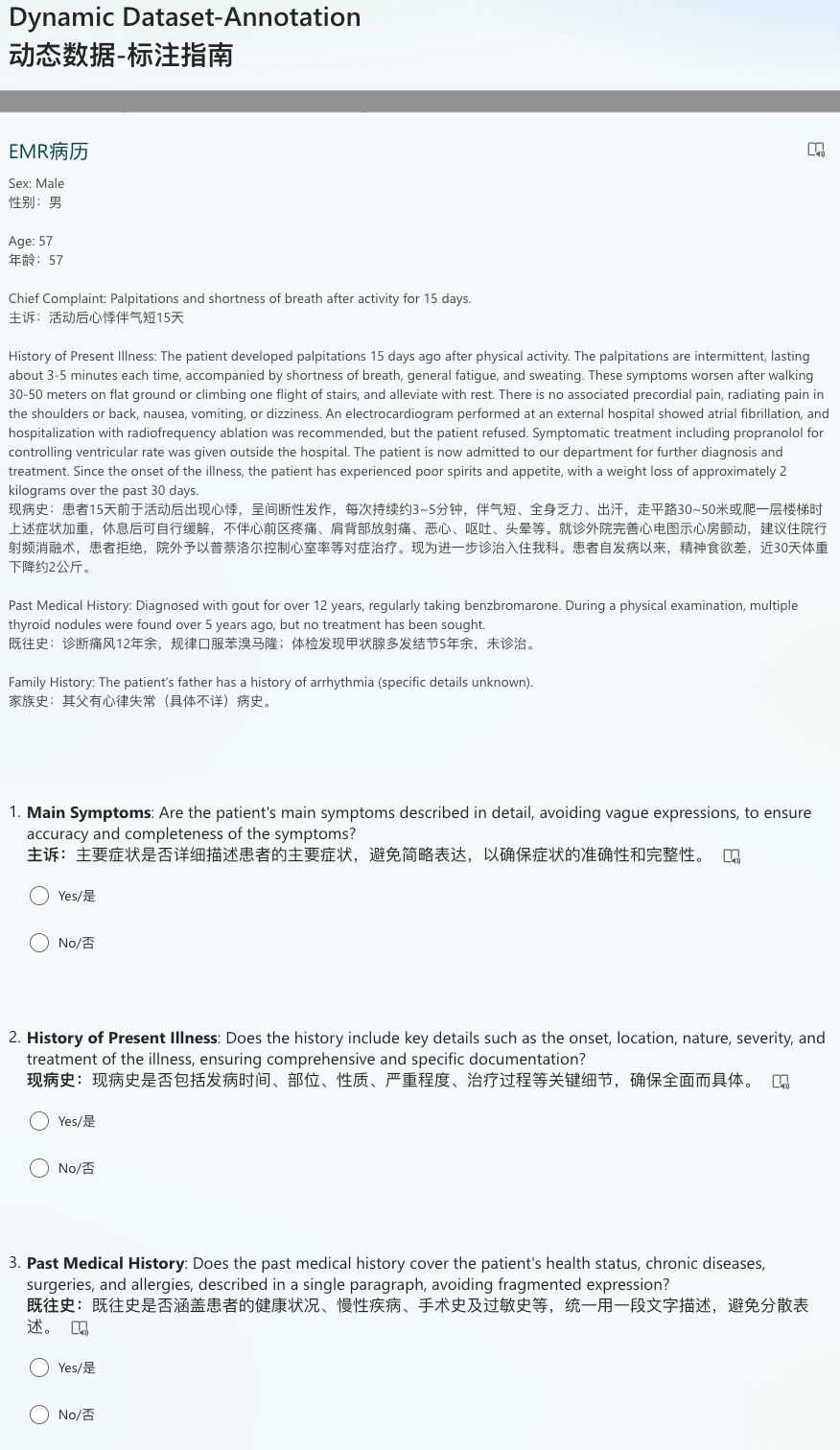}}
\caption{Dynamic dataset annotation guidance (continued)
} 
\label{fig: dynamic-annotation-guid}
\end{figure*}

\begin{figure*}[htbp]
\centerline{\includegraphics[width=1\textwidth]{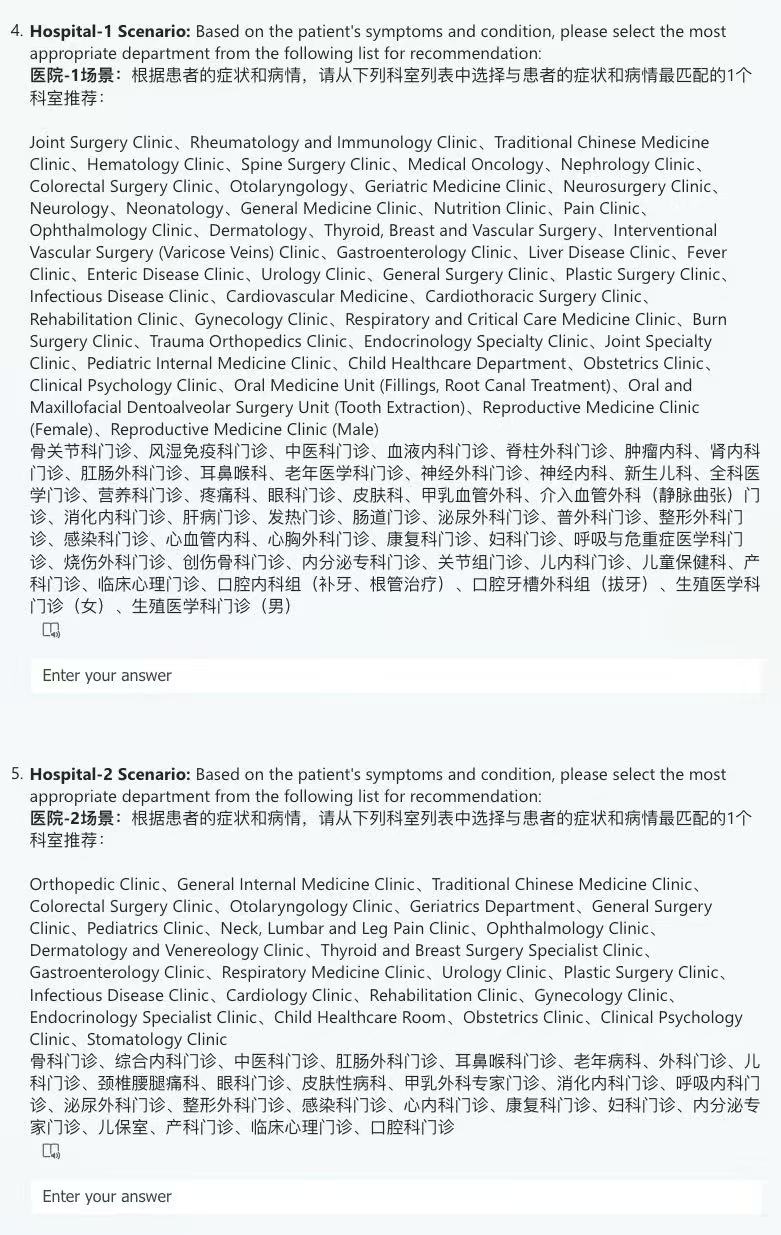}}
\caption{Dynamic dataset annotation guidance 
} 
\label{fig: dynamic-annotation-guid}
\end{figure*}

\section{OR-Dynamic Data Curation Process}
\label{apdx:dynamic_curation}

\noindent \textbf{Step 1: Establishing Selection Criteria.}
A senior medical expert reviewed 100 randomly chosen records from the raw corpus and defined three inclusion dimensions: \textit{Relevance} (case suitable for outpatient referral), \textit{Completeness} (sufficient symptom and context information), and \textit{Quality} (language coherence and medical plausibility). These criteria are detailed in Appendix~\ref{apdx:static_criteria}.

\noindent \textbf{Step 2: Automated Sampling.}
Following the established criteria, GPT-4o was prompted to automatically extract 2,000 candidate records from the de-identified corpus. The prompt constrained selection to cases satisfying all three inclusion dimensions.

\noindent \textbf{Step 3: Expert Annotation.}
Two licensed physicians—both experienced in outpatient triage—independently reviewed and labeled the 2,000 records according to department taxonomies provided by \textit{Hospital-1} (tertiary) and \textit{Hospital-2} (secondary). Disagreements were resolved by exclusion rather than arbitration to ensure high inter-annotator reliability.

\noindent \textbf{Step 4: Final Dataset.}
A total of 860 records with identical department labels assigned by both physicians were retained as the static evaluation dataset.

\section{Department Distribution}
\label{sec:department_distribution}

\begin{figure*}[htbp]
\centerline{\includegraphics[width=1\textwidth]{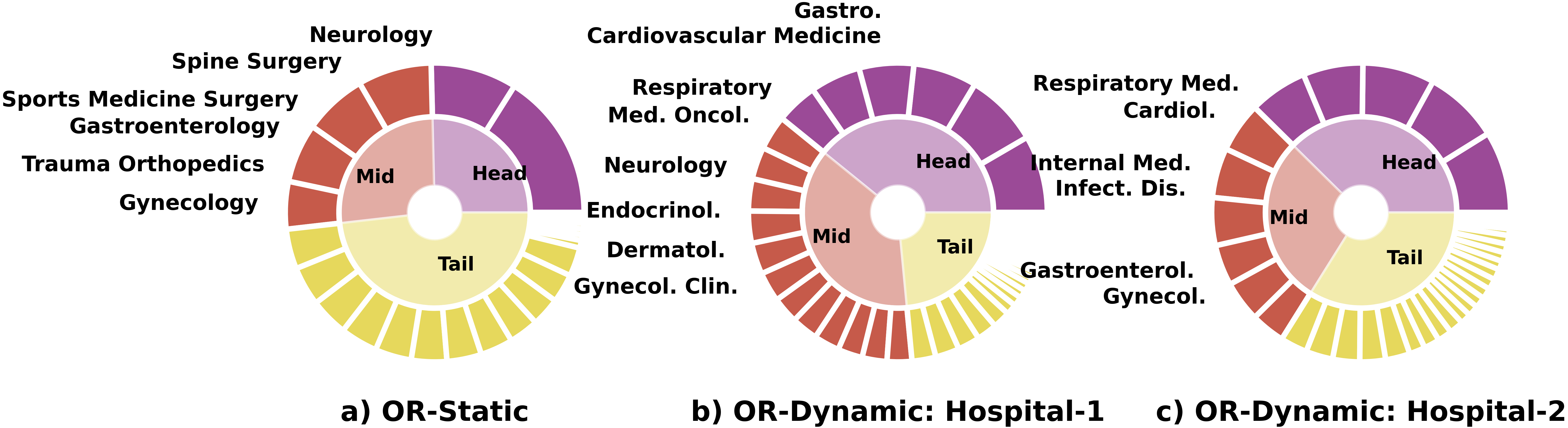}}
\caption{Distribution of department frequencies across the three datasets, shown with a two-level radial chart based on head/tail breaks. The size of each inner-ring segment reflects the overall proportion of the corresponding frequency group, while the outer ring further decomposes that group into department distribution.} 
\label{fig:or-static-sunburst}
\end{figure*}

\begin{figure*}[htbp]
\centerline{\includegraphics[width=1\textwidth]{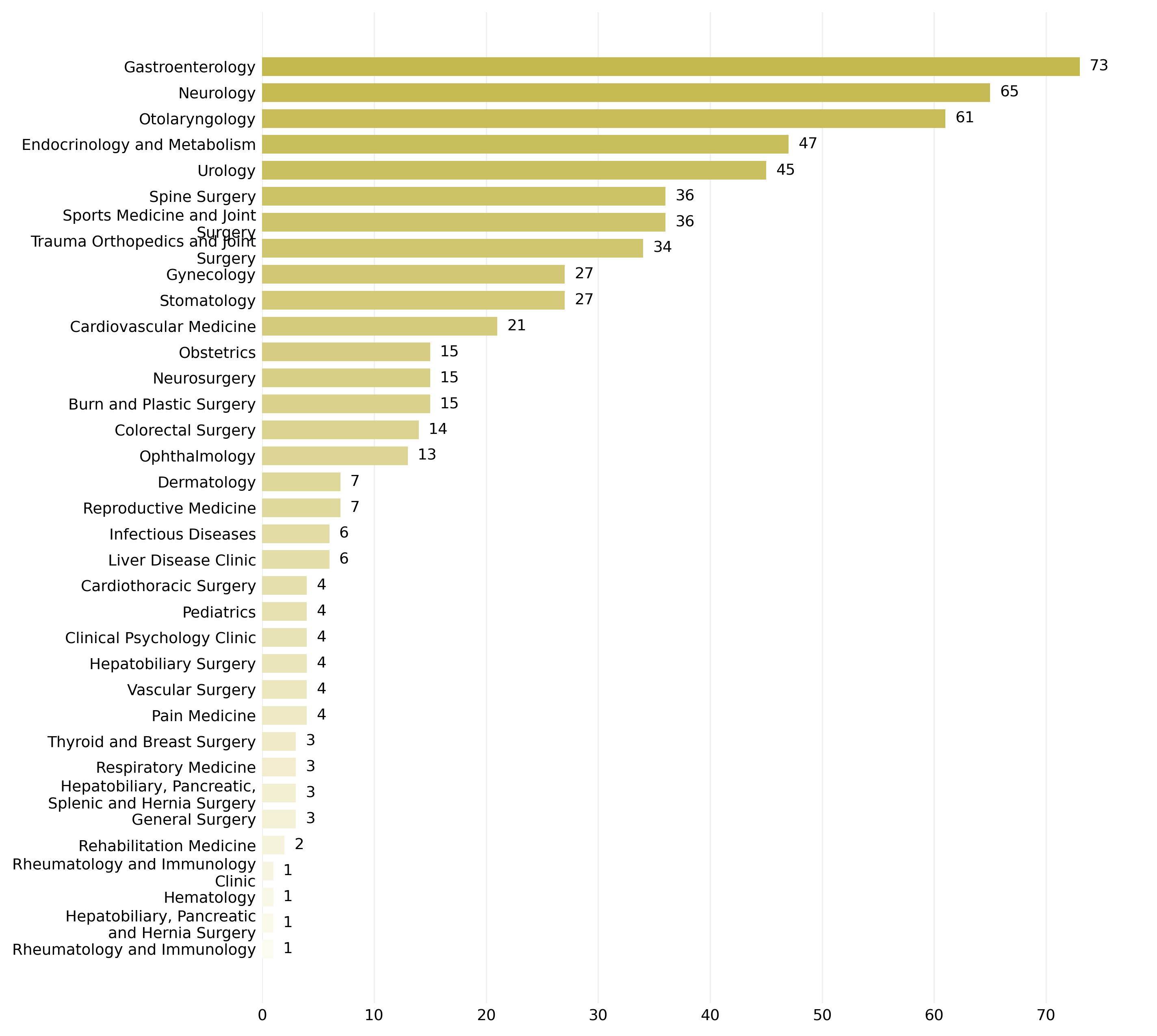}}
\caption{The distribution bar chart of the OR-static data. } 
\label{fig:or-static-barchart}
\end{figure*}

\begin{figure*}[htbp]
\centerline{\includegraphics[width=1\textwidth]{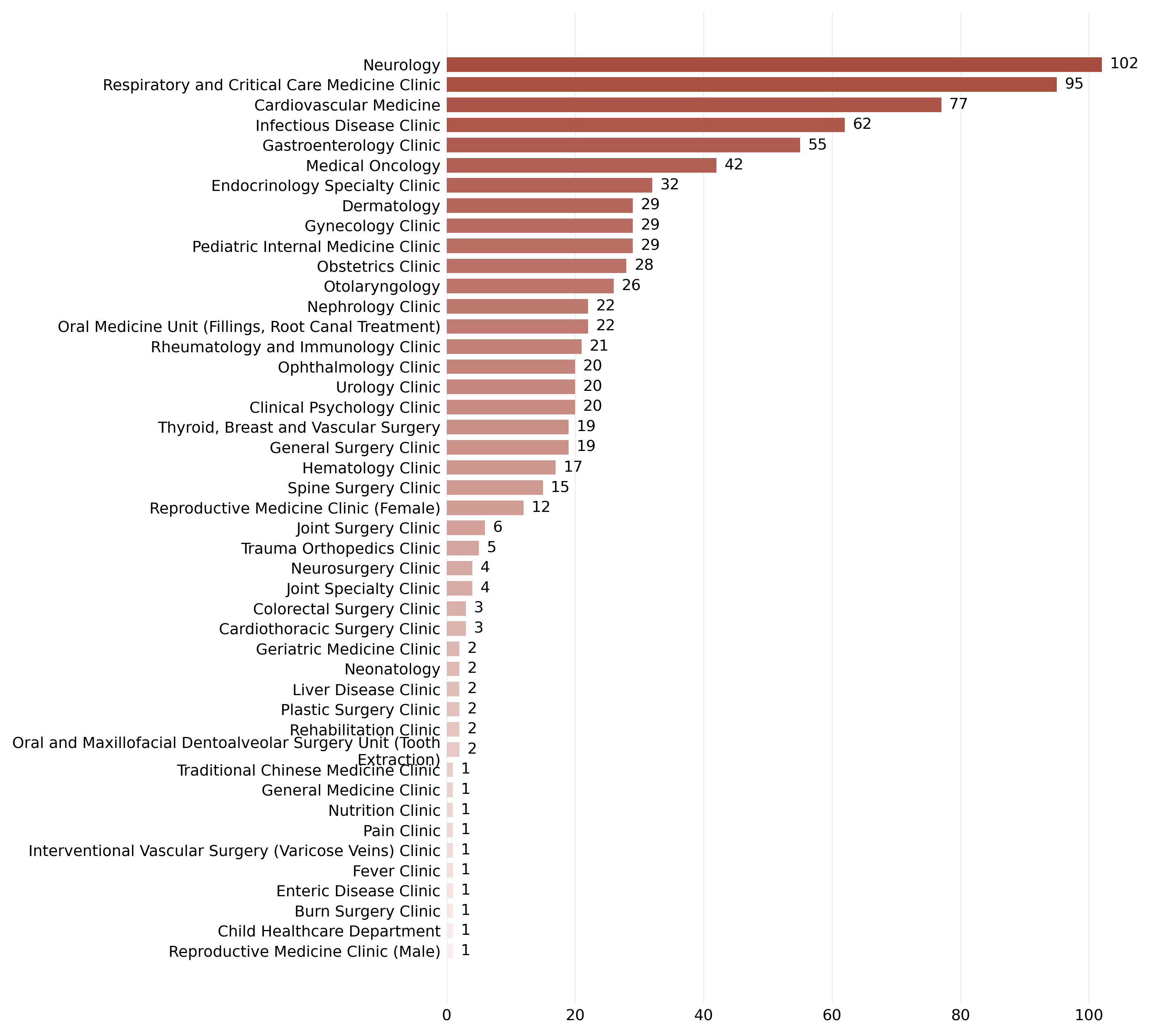}}
\caption{The distribution bar chart of the OR-dynamic data (hospital 1). } 
\label{fig:or-dynamic-barchart-h1}
\end{figure*}

\begin{figure*}[htbp]
\centerline{\includegraphics[width=1\textwidth]{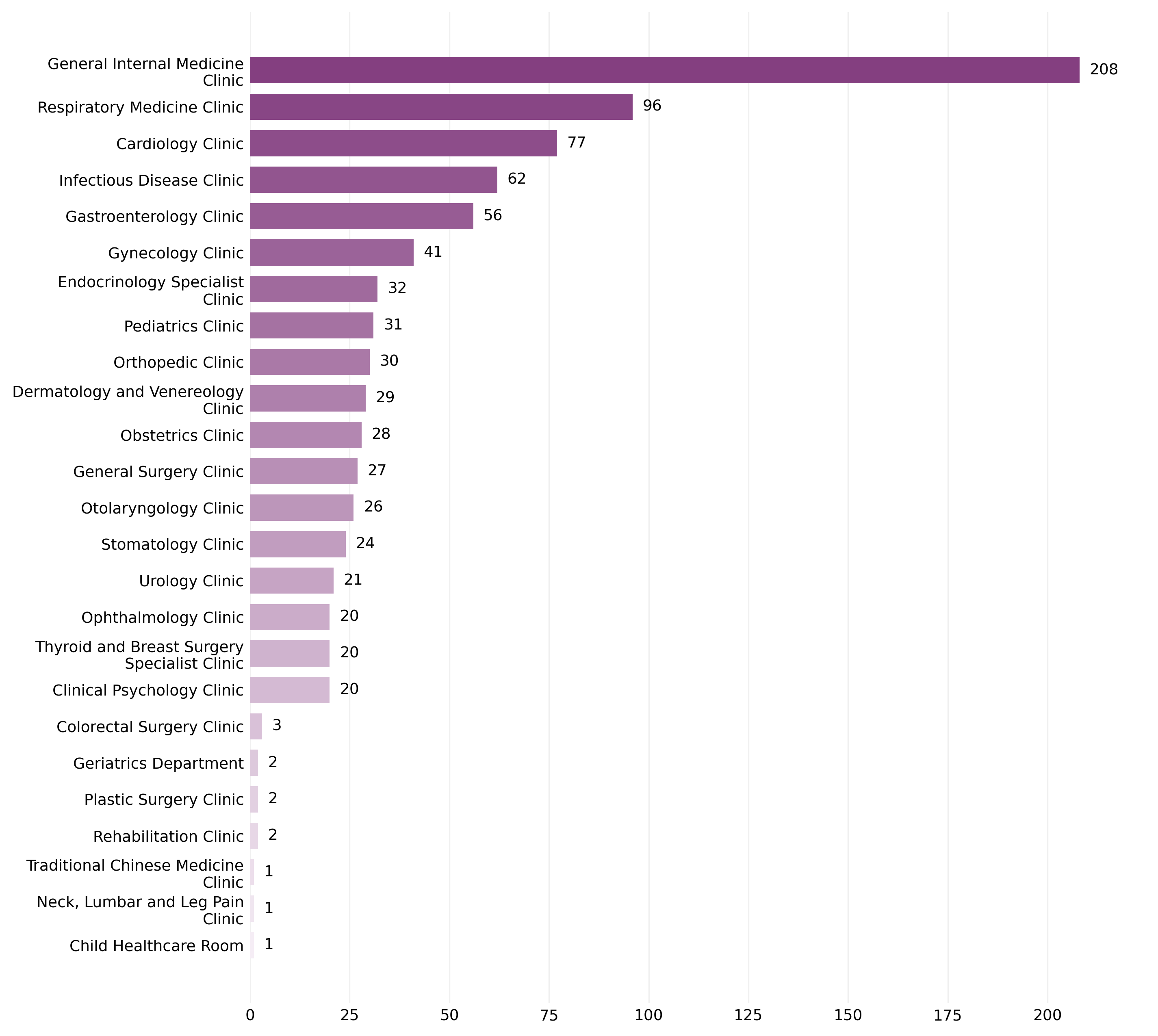}}
\caption{The distribution bar chart of the OR-dynamic data (hospital 2).} 
\label{fig:or-dynamic-barchart-h2}
\end{figure*}

To better characterize the label space of the outpatient referral task, we analyze the department distributions of OR-Static, OR-Dynamic Hospital-1, and OR-Dynamic Hospital-2. Figures~\ref{fig:or-static-barchart}, \ref{fig:or-dynamic-barchart-h1}, and \ref{fig:or-dynamic-barchart-h2} present the full frequency distributions of departments in each dataset, while Figure~\ref{fig:or-dynamic-barchart-h1} provides an additional head/tail breaks~\cite{Jiang_2013} view to highlight the overall concentration pattern. Since the department taxonomies are not fully aligned across datasets, our goal here is not to compare departments one by one, but to examine their shared and differing distributional properties, including head concentration, tail sparsity, category richness, and potential variation in label granularity.

All three datasets exhibit a clear \textbf{head-concentrated distribution}, as shown by the bar charts in Figures~\ref{fig:or-static-barchart}, \ref{fig:or-dynamic-barchart-h1}, and \ref{fig:or-dynamic-barchart-h2}, as well as the head/tail breaks visualization in Figure~\ref{fig:or-static-sunburst}. In each dataset, a small number of departments account for a disproportionately large share of cases, indicating that the referral distribution is dominated by a limited set of high-frequency specialties rather than being evenly spread across departments.

At the same time, all three datasets contain a substantial proportion of \textbf{extremely low-frequency categories}, with many departments appearing only a few times. This pattern is especially visible in the tail regions of Figure~\ref{fig:or-static-sunburst}, where a large number of departments occupy only very small segments. Such a sparse tail suggests that the referral task is not only long-tailed, but also affected by severe data scarcity for many rare specialties.

The two dynamic datasets further show a clear difference in \textbf{category richness}, with OR-Dynamic Hospital-1 covering noticeably more department types than OR-Dynamic Hospital-2. This implies that the two dynamic settings differ in task-space complexity: Hospital-1 requires the model to discriminate among a broader and potentially more fine-grained set of candidate departments, whereas Hospital-2 is relatively narrower in label space.

Among the three datasets, the \textbf{imbalance between top-ranked and lower-ranked categories} is most pronounced in OR-Dynamic Hospital-2. As shown in Figure~\ref{fig:or-dynamic-barchart-h2}, the frequency drops sharply from the leading department to the remaining categories, indicating a particularly steep rank--frequency curve. This suggests that Hospital-2 is more strongly dominated by a small number of common referral destinations, with the remaining departments forming a much thinner long tail.

Finally, the three datasets may not be fully aligned in \textbf{label granularity and head-category composition}. Some department labels are broad specialty names, whereas others correspond to more specific clinics or subspecialty services, and the dominant high-frequency departments also differ across datasets. Therefore, the comparison here focuses on distributional characteristics rather than exact one-to-one department correspondence, and these differences should be considered when interpreting cross-dataset referral difficulty.

\section{Model Selection Criteria}
\label{apdx:model-selection}

To ensure a comprehensive and fair evaluation across different model families, we selected models based on three criteria:
\begin{itemize}
    \item \textbf{Language alignment}: covering both English-centric and Chinese-centric models to reflect linguistic compatibility with the Chinese OR dataset.
    \item \textbf{Domain specialization}: including both general-purpose and medical-specific LLMs to evaluate the benefit of medical adaptation.
    \item \textbf{Architectural diversity}: covering encoder-based (e.g., BERT) and decoder-only (e.g., GPT-4o, Qwen2.5) across a wide range of model sizes (0.5B--671B).
\end{itemize}

This selection strategy enables systematic comparisons of scalability, generalization, and task suitability across both static and dynamic OR tasks. It also allows us to examine how different model characteristics—such as parameter scale, training objectives, and domain alignment—influence performance under varying supervision settings (e.g., fine-tuning vs. in-context learning). In particular, evaluating medical-specialized versus general-purpose models sheds light on the trade-off between broad generalization and clinical accuracy, while architectural comparisons reveal which design choices better support interactive reasoning in healthcare dialogues.

The LLMs include Gemma 9B-Instruct~\cite{FarabetWarkentin2024Gemma2}, Mixtral-8x22b-Instruct~\cite{jiang2024mixtral}, Apollo-72B~\cite{wang2024apollo}, Llama-3.2-90B-Vision-Instruct~\cite{dubey2024llama}, GPT-3.5~\cite{openai2023a}, GPT-4o~\cite{openai2023b}, 
o1-mini~\cite{openai_o1_2024}, o1~\cite{openai_o1_2024}, 
Claude 3.5~\cite{anthropic2024sonnet}, Gemini-Pro-1.5~\cite{team2023gemini}, 
BianQue-2~\cite{chen2023bianque},
Disc-MedLLM-13B~\cite{bao2023disc}, Yi-1.5-34B-Chat~\cite{ai2024yi}, DeepSeek-2.5-Chat~\cite{deepseekv2}, DeepSeek-v3~\cite{DeepSeekAI2025DeepSeekR1IR}, DeepSeek-r1~\cite{DeepSeekAI2025DeepSeekR1IR}, Moonshot-v1-128k~\cite{team2025kimi}, and the Qwen2.5 family of models~\cite{qwen2}. 

All evaluations were conducted using either API calls or local deployment. Fine-tuning and inference experiments were performed on 8 NVIDIA A40 GPUs, using a learning rate of 2e-5, a batch size of 16, and early stopping based on validation accuracy. We fix the random seed to 42 and set the decoding temperature to 0.0 for deterministic evaluation.

\section{Macro-F1 Score}~\label{apdx:staic_macrof1}
To complement accuracy, we compute Macro-F1 for a subset of models and static evaluation settings where per-department predictions are available.
Specifically, for each department, we compute the F1 score based on its precision and recall derived from the confusion matrix, and then report Macro-F1 as the unweighted average of F1 scores across all departments.
This formulation assigns equal importance to each department and is therefore less sensitive to class imbalance than overall accuracy. 

Comparing accuracy and Macro-F1 reveals notable disparities in class-wise performance that are not captured by accuracy alone.
From Table~\ref{tab:staticEvaluation_macroF1}, while several models achieve competitive accuracy, their Macro-F1 scores are substantially lower, indicating uneven performance across departments and systematic bias toward high-frequency classes.
In contrast, stronger models maintain more consistent Macro-F1 across prompting strategies, suggesting better balanced referral behavior rather than reliance on dominant departments.

\begin{table*}[h]
\centering
\small
\setlength{\tabcolsep}{10pt}
\resizebox{\textwidth}{!}{
\begin{tabular}{l|r|rrrr|l}
\toprule
\multicolumn{1}{l|}{\textbf{Models}}  
& \multicolumn{1}{c|}{\textbf{\#Params (B)}}  
& \multicolumn{1}{c|}{\textbf{Zero Shot$\uparrow$ }} 
& \multicolumn{1}{c|}{\textbf{Few Shot$\uparrow$}} 
& \multicolumn{1}{c|}{\textbf{Majority Vote$\uparrow$}} 
& \multicolumn{1}{c|}{\textbf{CoT$\uparrow$}} 
& \multicolumn{1}{c}{\textbf{Avg.}} \\ 
\midrule

\rowcolor{gray!10} \multicolumn{7}{c}{\textit{Medical LLMs}} \\ \midrule
\textbf{BianQue-2} & 6.5B 
& 13.66 & 6.25 & 12.36 & 6.04 & 9.58 \\

\textbf{DISC-MedLLM} & 13B 
& 27.14 & 26.69 & 33.53 & 31.64 & 29.75 \\

\midrule
\rowcolor{gray!10} \multicolumn{7}{c}{\textit{English-centric LLMs}} \\ \midrule

\textbf{Gemma-Instruct} & 9B 
& 42.32 & 45.56 & 45.02 & 45.97 & 44.72 \\

\textbf{Llama-3.2-Instruct} & 90B 
& 43.64 & 43.95 & 46.73 & 42.75 & 44.27 \\

\textbf{GPT-4o} & - 
& 40.44 & 49.35 & 42.68 & 45.26 & 44.43 \\

\textbf{o1} & - 
& \textbf{57.97} & \textbf{57.61} & \textbf{55.53} & \textbf{55.70} & \textbf{56.70} \\

\bottomrule
\end{tabular}
}
\caption{\textbf{Static Evaluation} performance (Macro-F1 \%) across different prompting strategies. Avg.\ denotes the average Macro-F1 score. ``--'' indicates models with undisclosed parameter counts.}
\label{tab:staticEvaluation_macroF1}
\end{table*}

\section{Model Fine-tuning Experiments} \label{apdx:bert_finetuning}

To comprehensively compare modeling paradigms for the OR-static task under data-constrained settings, we selected three representative models that reflect distinct architectural families and learning strategies:
\begin{itemize}
    \item \textbf{Bert-Base-Chinese~\cite{fu2024monarch}}: a \textit{discriminative encoder-based model} representing traditional \textbf{classification} architectures trained via supervised fine-tuning (SFT);
    \item \textbf{Qwen2.5-3B}: a \textit{decoder-only causal language model}, standing in for \textbf{generative} architectures that also undergo supervised fine-tuning using instructional prompts;
    \item \textbf{GPT-4o}: a large-scale proprietary \textit{decoder-only model} used under a \textbf{few-shot in-context learning} setting, representing the zero-parameter-update paradigm typical for frontier LLMs that cannot be fine-tuned directly.
\end{itemize}
These three representative models analysis from two key perspectives: (1) \textbf{architecture type}—encoder-based (Bert-Base-Chinese) vs. decoder-only (Qwen2.5-3B, GPT-4o); and (2) \textbf{training paradigm}—parameter-updating (SFT) vs. inference-time conditioning (in-context learning).

All experiments were conducted on the same OR-Static dataset, consisting of 612 patient-AI dialogues annotated with department labels. The dataset was split into 80\% for training (490 samples), 10\% for validation (61 samples), and 10\% for testing (122 samples). To simulate low-resource scenarios, we subsampled the training data into 7 subsets of increasing size: 50, 100, 150, 200, 250, 300, and 400 samples. Each model was evaluated on the same held-out test set.

\noindent \textbf{Supervised Fine-Tuning (BERT and Qwen2.5-3B)}~ 
For Bert-Base-Chinese, \textit{inputs} were tokenized dialogue texts, and the \textit{output} was a softmax classification over department labels. Fine-tuning was performed with a learning rate of 2e-5, batch size of 16, and for 3 epochs, with early stopping based on validation accuracy.

For Qwen2.5-3B, we adopted prompt-based training using a fixed instruction appended to the dialogue history, followed by the correct department name as the target sequence. We used a constant temperature of 0.0 for decoding and the same optimizer settings as BERT. Both models were trained on two NVIDIA A40 GPUs.

\noindent \textbf{Few-Shot In-Context Learning (GPT-4o).}~ 
GPT-4o was evaluated without parameter updates. For each test case, we constructed prompts as formatted in Appendix~\ref{apdx: static-prompt-design}. Since GPT-4o does not require training infrastructure, all inference was conducted via API access under consistent prompt templates.

This experimental design allows us to systematically compare performance across models with different inductive biases, sizes, and learning modalities—providing insight into sample efficiency, saturation behavior, and the trade-offs between fine-tuning and in-context learning in clinical classification tasks.

\section{Language-Centric Analysis}\label{sec:language-analysis}

In this section, we present a detailed comparison of language-centric model performance, as classified into \textit{English-centric} and \textit{Chinese-centric} categories in Table~\ref{tab: staticEvaluation} and Table~\ref{tab: dynamicResult}. The classification is based on each model’s primary training language and design orientation. This analysis aims to uncover whether linguistic alignment with the task language (Chinese) influences model effectiveness in static classification and dynamic dialogue settings.

\noindent \textbf{Static Evaluation Comparison}

As shown in Table~\ref{tab: staticEvaluation}, Chinese-centric LLMs generally outperform their English-centric counterparts in static classification. For instance, models such as DeepSeek-v3 and Qwen2.5-Instruct (both Chinese-centric) achieve average static accuracies exceeding 78\%, while most English-centric models such as GPT-3.5, Claude 3.5, and LLaMA-3.2-Instruct remain in the 65–75\% range. The best-performing English-centric model (o1) achieves an average of 77.9\%, closely rivaling Qwen2.5-Instruct (78.19\%) and DeepSeek-v3 (80.27\%).

This suggests that although English-centric models benefit from stronger general capabilities due to larger scale or broader pretraining corpora, Chinese-centric models possess a natural linguistic alignment advantage when applied to Chinese outpatient referral tasks. Moreover, the prompt sensitivity observed in static evaluations (Section~\ref{sec:static-evaluation-setup}) appears to affect English-centric models more severely, possibly due to less exposure to structured medical queries in Chinese. This performance gap highlights the language-adaptive pretraining and domain-specific instruction tuning when deploying LLMs in non-English clinical contexts.

\noindent \textbf{Dynamic Evaluation Comparison}

In the dynamic scenario (Table~\ref{tab: dynamicResult}), the distinction between language-centric groups becomes more nuanced. While Chinese-centric models like DeepSeek-v3 and Qwen2.5 still perform competitively—especially on Hospital-2’s simpler classification task—English-centric models such as GPT-4o and o1 also demonstrate robust performance, achieving over 75\% accuracy across multiple dialogue turns.

Interestingly, GPT-4o shows consistent accuracy improvement with each round of dialogue (Turn 0 to Turn 5), indicating its strong capability to handle information accumulation and iterative reasoning, despite its English-centric training background. In contrast, some smaller Chinese-centric models, such as Yi-1.5 or early Qwen versions (e.g., 0.5B), experience early saturation or degradation across turns, revealing challenges in long-context understanding and dialogue coherence.

This observation suggests that while linguistic alignment offers benefits in static, classification-style settings, success in dynamic, interactive tasks may rely more on reasoning capabilities, training architecture, and instruction tuning quality rather than language alignment alone.

These findings highlight the importance of balancing linguistic alignment with architectural strength and training diversity when deploying LLMs for outpatient referral tasks in specific language settings such as Chinese.

\section{Prompt Sensitivity Analysis in Static Evaluation}
\label{apdx:prompt-analysis}

Table~\ref{tab: staticEvaluation} presents the performance of various LLMs on the OR task under four prompting strategies: zero-shot, few-shot, chain-of-thought (CoT), and majority vote. 

We observe that model performance varies significantly with prompt choice, especially among smaller or domain-specific models. For instance, medical-specialized LLMs such as DISC-MedLLM show sharp performance drops under few-shot and CoT prompting, indicating limited generalization and higher prompt sensitivity. In contrast, general-purpose large models such as DeepSeek-v3, GPT-4o, and o1 maintain high and stable performance across all prompting strategies. Notably, DeepSeek-v3 achieves the highest CoT accuracy (80.88\%) while also performing strongly in few-shot (80.39\%) and majority vote (80.23\%), demonstrating reasoning adaptability.

While zero-shot prompting provides a baseline for the OR task, it tends to underperform in most models compared to tailored strategies. In particular, few-shot prompting, which provides several the OR examples, yields large improvements for models like GPT-3.5, but induces volatility in smaller models such as Yi-1.5-Chat, whose few-shot accuracy drops below 45\%. This suggests that few-shot effectiveness is highly dependent on model scale.

we also notice clear variation within the same model across the four prompting settings. For example, Claude-3.5-Sonnet ranges from 77.47\% (zero shot), 77,21 (majority vote) to 70.84\% (few shot) and 66.67\% (CoT). Yi-1.5-Chat drops from 70.75\% (majority vote) to 43.99\% under different prompts. In contrast, models like DeepSeek-v3, GPT-4o, and o1 show smaller gaps, indicating more stable performances. These differences highlight that model behavior can shift significantly with prompt format, making careful prompt design essential for reliability in clinical use.

Overall, these results confirm that prompt sensitivity remains a critical bottleneck in applying LLMs to real-world clinical classification tasks. Models with higher parameter counts and broader pretraining demonstrate better prompt-agnostic behavior, whereas smaller or domain-restricted models exhibit significant variance, underlining the need for targeted prompt calibration when deploying LLMs in medical settings.

\section{Hospital-Specific Analysis in Dynamic Evaluation}
\label{apdx:hospital-analysis}

To better understand how department complexity and hospital characteristics influence model performance, we compare dynamic evaluation results across Hospital-1 and Hospital-2 (as described in Section~\ref{sec: Dynamic_eval_data}). Hospital-1 (tertiary hospital) involves 45 departments, presenting a more fine-grained classification task. In contrast, Hospital-2 (secondary hospital) has only 25 departments, posing a relatively simpler challenge.
(\textit{A tertiary hospital provides highly specialized medical care with advanced equipment and expert staff, while a secondary hospital offers general specialist services for referred patients but with less complexity and capacity.})

From Table~\ref{tab: dynamicResult}, we observe that most models, including general-purpose LLMs such as GPT-4o and Qwen2.5, perform consistently better on Hospital-2 than Hospital-1. This gap is especially pronounced in earlier dialogue turns (0–2), suggesting that reduced departmental granularity leads to faster convergence and higher initial accuracy. Interestingly, high-capacity models like DeepSeek-v3 and Moonshot narrow this gap in later turns through effective questioning, while smaller models struggle to maintain performance consistency.

These findings highlight the importance of adapting referral systems to hospital-specific configurations, particularly in terms of department taxonomy size and distribution complexity.

\noindent \textbf{Limited Benefit of \textit{Medical} LLMs over \textit{General} LLMs}~ 
Medical models like DISC-MedLLM are expected to perform better in medical tasks. However, in static evaluation (Table~\ref{tab: staticEvaluation}), GPT-4o and DeepSeek-v3 achieved higher accuracy. This may be because general-purpose LLMs may outperform medical LLMs because their frequent updates incorporate newer knowledge. In contrast, medical LLMs often rely on outdated architectures or data."

\begin{takeaway}
    Current medical LLMs show little advantage over general-purpose LLMs.
\end{takeaway}

\section{Static-Evaluation Prompt Design}\label{apdx: static-prompt-design}

The static evaluation framework assesses the model's ability to recommend hospital departments. Prompt sensitivity is a widely observation in LLMs, where slight changes may lead to different performance~\cite{zhao2021calibrate, wei2022chain, zhuo-etal-2024-prosa, habba-etal-2025-dove}. Accordingly, we incorporate four prompting strategies into the evaluation: zero-shot prompting, few-shot prompting, chain-of-thought (CoT) prompting, and a majority vote strategy.

\begin{CJK}{UTF8}{gkai}
\begin{figure*}[ht]
\centering
\begin{tcolorbox}[title=Zero-shot Prompt]
    \texttt{You are a doctor at the hospital's reception desk, responsible for directing patients to the appropriate department. The department you are currently working with is:
\{DEPARTMENT LIST\}.
Given the conversation history so far, you need to recommend the most suitable department from the hospital's directory. }
\vspace{0.5cm}
    \hrule  
    \vspace{0.5cm}
你是医院导诊台的工作人员，负责将患者引导到合适的科室。当前科室有：\texttt{\{DEPARTMENT LIST\}}。 请根据对话内容推荐最合适的科室。
\end{tcolorbox}
\caption{Zero-shot  Prompt}
\label{fig: zero-shot-prompt}
\end{figure*}

\begin{figure*}[ht]
\centering
\begin{tcolorbox}[title=Few-shot Prompt]
    \texttt{You are a doctor at the hospital's reception desk, responsible for directing patients to the appropriate department. The department you are currently working with is:
\{DEPARTMENT LIST\}.
Below are some example dialogues. Please refer to these examples and recommend the most suitable department based on the given conversation.}

\begin{itemize}
    \item \textbf{\texttt{Example 1}}

\textbf{\texttt{Conversation}}

\texttt{Patient: Hi, doctor, I’ve been feeling very tired lately and have been in low spirits.}

\texttt{Doctor: How long have you been feeling this way?}

\texttt{Patient: It's been about two weeks now.}

\textbf{\texttt{Recommendation}} 

\texttt{Clinical Psychology}
      \item \textbf{\texttt{Example 2}}

\textbf{\texttt{Conversation}}

\texttt{Patient: I’ve been coughing a lot recently and sometimes find it hard to breathe.}

\texttt{Doctor: Does the coughing worsen at any particular time of day?}

\texttt{Patient: It's worse at night, and I feel more shortness of breath then too.}

\textbf{\texttt{Recommendation}}

\texttt{Respiratory and Critical Care Medicine}

    \item \textbf{\texttt{Example 3}}
    
\textbf{\texttt{Conversation}}

\texttt{Patient: My child has been running a fever for the past few days and has a bit of a cough.}

\texttt{Doctor: Has the fever been consistent, or does it come and go?}

\texttt{Patient: It seems to come and go, but the cough has been constant.}

\textbf{\texttt{Recommendation}}

\texttt{Pediatrics}
\end{itemize}
\vspace{0.5cm}
    \hrule  
    \vspace{0.5cm}
你是医院接待处的医生，负责将患者引导到合适的科室。你负责的科室有：\{DEPARTMENT LIST\}。以下是一些对话示例。
\begin{itemize}
    \item 示例一 \\ \\
\textit{对话}\\
患者：医生，我最近总是感到非常疲劳，情绪也很低落。\\
医生：这种情况持续多久了？\\
患者：大约有两周了。\\
\textbf{推荐科室:} 临床心理科
\end{itemize}
\end{tcolorbox}
\caption{Few-shot Prompt (continued)}
\label{fig: zero-shot-prompt}
\end{figure*}

\begin{figure*}[ht]
\centering
\begin{tcolorbox}[title=Few-shot Prompt]
\begin{itemize}

    \item 示例二 \\ \\
\textit{对话} \\
患者：我最近常咳嗽，有时还觉得呼吸困难。\\
医生：咳嗽在特定时间有加重吗？\\
患者：晚上更严重，而且感觉呼吸更困难。\\

\textbf{推荐科室:} 呼吸与危重症医学科 \\

\item 示例三 \\ \\
\textit{对话} \\
患者：我的孩子这几天一直发烧，还咳嗽。\\
医生：发烧是持续的吗？还是时有时无？\\
患者：有时发烧，有时不发烧，但咳嗽一直没停。\\

\textbf{推荐科室:} 儿科 \\
    
\end{itemize}

\end{tcolorbox}
\caption{Few-shot  Prompt}
\label{fig: few-shot-prompt}
\end{figure*}

\begin{figure*}[ht]
\centering
\begin{tcolorbox}[title=CoT Prompt]
    \texttt{ You are a doctor at the hospital's reception desk, responsible for directing patients to the appropriate department. The department you are currently working with is:
\{DEPARTMENT LIST\}.
Given the conversation history so far, you need to recommend the most suitable department from the hospital's directory. Here are the reasoning steps you need to follow:
\begin{itemize}
    \item First, carefully read the conversation and extract the main symptoms described by the patient. Please explain in detail why you consider these to be the primary symptoms.
    \item Next, based on the extracted symptoms, match them with the most suitable department from the provided list. Please describe the matching process for each department and explain why you ultimately chose a specific department.
    \item Finally, based on your reasoning process, summarize and explain why this department is the most appropriate choice for the patient's current situation.
\end{itemize}}

\end{tcolorbox}
\caption{CoT Prompt (Continued)}
\label{fig: cot-prompt}
\end{figure*}

\begin{figure*}[ht]
\centering
\begin{tcolorbox}[title=CoT Prompt]
你是医院导诊台的医生，负责将患者引导到合适的科室。当前负责的科室有：\{DEPARTMENT LIST\}。根据目前的对话记录，请按以下步骤推荐最适合的科室：
\begin{itemize}
    \item 首先，仔细阅读对话内容，提取患者描述的主要症状，并详细说明你为什么认为这些症状是主要症状。

\item 接着，根据提取的症状，从提供的科室列表中匹配最合适的科室。请描述匹配过程并解释你最终选择该科室的理由。

\item 最后，基于你的推理过程，总结并解释为什么该科室是患者当前情况最合适的选择。
\end{itemize}
\end{tcolorbox}
\caption{CoT Prompt}
\label{fig: cot-prompt}
\end{figure*}

\end{CJK}
The goal of the experiment of the static evaluation is to assess whether the findings are sensitive to prompting engineering. Previous work has demonstrated that evaluation results can be significantly influenced by prompting strategies. For instance, existing studies have shown how prompting can impact LLM performance~\cite{zhao2021calibrate, wei2022chain, zhuo-etal-2024-prosa, habba-etal-2025-dove}. This underscores the importance of careful prompt design in evaluating model performance, validating our approach to testing the sensitivity of our findings to different prompt engineering strategies.

\noindent \textbf{Zero-Shot} In the zero-shot setting, the model is prompted without any specific examples provided in advance~\cite{radford2019language}. The prompt simply instructs the model to recommend a department based on the patient's description. This approach tests the model’s inherent ability to generalize from its pre-training to the task at hand. See Figure \ref{fig: zero-shot-prompt}.

\noindent \textbf{Few-Shot}This method involves providing the model with a small number of example dialogues, each paired with their correct department classifications, before testing it on new dialogues~\cite{brown2020language}. The few-shot approach tests the model's ability to adapt and learn from limited data. It offers insights into the model's flexibility and efficiency in incorporating new information into its decision-making process, which is crucial for tasks that involve sparse or rapidly evolving data. See Figure \ref{fig: few-shot-prompt}.

\noindent \textbf{CoT}This approach prompts the model to generate a detailed reasoning process before arriving at a final department prediction. The CoT method~\cite{wei2022chain} is instrumental in evaluating the model's ability to articulate its thought process, providing a window into how the model reasons through the information provided. By analyzing these detailed responses, we can assess the model's logical consistency, depth of understanding, and potential biases in reasoning. See Figure\ref{fig: cot-prompt}. 

\noindent \textbf{Majority Vote}This technique involves aggregating multiple predictions from different model configurations or runs, with the final output determined by the department receiving the most votes~\cite{yang2024llm}. The majority vote method assesses the consistency and stability of the model's predictions across different instances. It helps in understanding how variations in model setup or random factors might influence the outcomes, thereby providing a measure of the reliability of the model's performance.

\textbf{\textit{How can CoT be considered a more complex task than zero-shot?}} \ 
Zero-shot prompting typically leads the model to make an immediate prediction based on the input, without requiring it to justify or elaborate on its decision. In contrast, Chain-of-Thought (CoT) prompting guides the model to produce intermediate reasoning steps before arriving at a final answer. This makes CoT inherently more demanding, especially in the context of outpatient referral, where cases often involve overlapping symptoms, ambiguous histories, and multiple plausible departments. Instead of guessing, the model must synthesize information, weigh differential diagnoses, and construct a coherent rationale—closely mirroring the reasoning required in real-world clinical settings.

\section{Dynamic-Evaluation Prompt Design} \label{apdx: dynamic-prompt-design}
For the patient agent prompt design, see Figure~\ref{fig: patient-agent-prompt}.
For the Questioning Doctor agent prompt design, see Figure~\ref{fig: question-doc-prompt}.
For the OR Doctor agent prompt design, see Figure~\ref{fig: ior-doctor-prompt}. For the Department selection doctor agent prompt design, see Figure~\ref{fig:ior-doctor-department-prompt}
\begin{CJK}{UTF8}{gkai}
\begin{figure*}[ht]
\centering
\begin{tcolorbox}[title=Patient-Agent Prompt]
    \texttt{ You are \{ROLE\},. Your task is to communicate with the doctor and respond to their questions based on their inquiries. Here is your information: \\1. Chief Complaint: \{CHIEF\_COMPLAINT\} \\2. History of Present Illness: \{HISTORY\_OF\_PRESENT\_ILLNESS\} \\3. Past Medical History: \{PAST\_MEDICAL\_HISTORY\} \\4. Family History: \{FAMILY\_HISTORY\}.}

\vspace{0.5cm}
    \hrule  
    \vspace{0.5cm}

    你是 \{ROLE\}，你的任务是根据医生的询问与其沟通并作出回应。以下是你的信息：

1. 主诉：\{CHIEF\_COMPLAINT\} \\
2. 现病史：\{HISTORY\_OF\_PRESENT\_ILLNESS\} \\
3. 既往病史：\{PAST\_MEDICAL\_HISTORY\} \\
4. 家族史：\{FAMILY\_HISTORY\}

\end{tcolorbox}
\caption{Patient-Agent Prompt}
\label{fig: patient-agent-prompt}
\end{figure*}

\newpage
\begin{figure*}[ht]
\centering
\begin{tcolorbox}[title=Questioning-Doctor-Agent Prompt]
    \texttt{ You are a doctor asking questions during the outpatient referral process. In order to guide the patient to provide as much relevant information about their symptoms as possible, you should continuously ask key questions related to department recommendations.}

     \vspace{0.5cm}
    \hrule  
    \vspace{0.5cm}
你是一位负责提问的医生，在导诊过程中引导患者提供症状信息。为了让患者详细描述症状并确保导诊的准确性，你需要根据患者的回答不断提问，确保获取尽可能多的相关信息，从而为患者推荐最合适的科室。
\end{tcolorbox}
\caption{Questioning-Doctor-Agent Prompt}
\label{fig: question-doc-prompt}
\end{figure*}

\begin{figure*}[ht]
\centering
\begin{tcolorbox}[title=IOR-Doctor-Agent Prompt]
    \texttt{ You are an outpatient referral doctor at the hospital’s front desk, responsible for recommending the appropriate department to the patient. The department you are in charge of is: \{DEPARTMENT\}. Based on all the dialogue information between you and the patient, you should carefully consider which department the patient may need to visit, and recommend a department from the department list. The department name must match exactly, with no variations.}

    \vspace{0.5cm}
    \hrule  
    \vspace{0.5cm}

你是一位导诊医生，负责根据患者的症状推荐合适的科室。你当前负责的科室是：\{DEPARTMENT\}。根据你与患者之间的所有对话信息，你要仔细考虑患者可能就诊的科室，并从科室列表中推荐最合适的科室。科室名称必须完全匹配，不能有任何变化。

\end{tcolorbox}
\caption{IOR-Doctor-Agent Prompt}
\label{fig: ior-doctor-prompt}
\end{figure*}

\begin{figure*}[ht]
\centering
\begin{tcolorbox}[title=IOR-Doctor-Department-Agent Prompt]
\textbf{English.}
You are the analysis assistant for outpatient referral. Based only on the current dialogue history, assign a \textbf{relative support score} to \textbf{each candidate department} over 7 evidence dimensions.

\textbf{Rules:}
(1) Use only observed dialogue information; if missing, assign a low score or 0.  
(2) Do not ask questions, reply to the patient, or recommend a department.  
(3) You must score \textbf{all} candidate departments.

\textbf{Dimensions:}
(1) Chief complaint match (0--3);
(2) Symptom consistency (0--3);
(3) Red flags / urgency relevance (0--3);
(4) Patient background fit (0--3);
(5) Tests / history / medication support (0--3);
(6) Department management fit (0--3);
(7) Contradictory evidence (0, $-1$, $-2$, $-3$).

\textbf{Output:}
For each department, provide: department name, 7 scores, \texttt{total\_support\_score}, and a one-sentence rationale.

\vspace{0.4cm}
\hrule
\vspace{0.4cm}

\textbf{中文.}
你是导诊医生的分析助手。只能依据当前对话信息，对\textbf{每个候选科室}按7个证据维度进行\textbf{相对支持度评分}。

\textbf{要求：}
（1）仅依据已出现的信息；若缺失，给低分或0分，不补充假设。  
（2）禁止提问、回复病人、推荐科室。  
（3）必须对\textbf{全部}候选科室评分。

\textbf{维度：}
（1）主诉匹配度（0--3）；
（2）症状一致性（0--3）；
（3）危险信号/紧急性（0--3）；
（4）人群与背景适配性（0--3）；
（5）检查/病史/用药支持度（0--3）；
（6）科室处置匹配度（0--3）；
（7）矛盾证据强度（0，$-1$，$-2$，$-3$）。

\textbf{输出：}
对每个科室输出：科室名称、7项分数、\texttt{total\_support\_score} 和一句 \texttt{rationale}。

\end{tcolorbox}
\caption{Prompt for the IOR-Doctor-Department-Agent.}
\label{fig:ior-doctor-department-prompt}
\end{figure*}

\end{CJK}.

\section{Fidelity Analysis of Patient Simulators}
\label{apdx:simulator_fidelity}
\noindent \textbf{Evaluation Setup}

To evaluate the fidelity of LLM-based patient simulators, we conduct a controlled comparison between human role-played patients and LLM-simulated patients. Specifically, we randomly select 30 cased from the OR-Dynamic dataset.

As a human patient baseline, three licensed physicians with clinical experience in outpatient referral and prior involvement in standardized patient--based training are invited to role-play patients.
For LLM-based patient simulators, we evaluate GPT-4o and DeepSeek-2.5-Chat.

Both human physicians and LLM-based simulators interact with the same doctor model (played by GPT-4o) under an identical dialogue protocol. The doctor model asks follow-up questions iteratively and produces a referral decision once sufficient information is obtained.

This fidelity analysis follows an independent evaluation protocol that focuses exclusively on patient-side interaction behavior and is analytically decoupled from the main OR-Dynamic experiments, ensuring that simulator validation does not bias doctor-side performance evaluation.

\noindent \textbf{Scoring Criteria}

We evaluated along four dimensions recommended by licensed physicians with outpatient referral experience:

\textbf{Dialogue Length}: Average number of words per patient response, reflecting verbosity and realism of symptom description.

\textbf{Response Confidence}: Expert-rated confidence level of patient responses on a 5-point Likert scale, assessing whether answers are decisive and appropriate given the inquiry.

\textbf{Logical Coherence}: Expert-rated coherence on a 5-point Likert scale, measuring internal consistency across turns and absence of contradictions.

\textbf{Interaction Efficiency}: Average number of dialogue turns required before a referral decision is made, reflecting efficiency of information disclosure.

These criteria focus on interactional plausibility rather than medical correctness, aligning with the simulator’s role as a controlled dialogue environment.

\begin{table*}[h]
\small
\resizebox{\textwidth}{!}{
\begin{tabular}{l|rrrrrr|rrrrrr}
\toprule & \multicolumn{6}{c|}{\textbf{Hospital-1}}                                    & \multicolumn{6}{c}{\textbf{Hospital-2}}                                     \\ \midrule
\textbf{Turn}                       & \textbf{0} & \textbf{1} & \textbf{2} & \textbf{3} & \textbf{4} & \textbf{5} & \textbf{0} & \textbf{1} & \textbf{2} & \textbf{3} & \textbf{4} & \textbf{5} \\ \midrule

\multicolumn{13}{c}{\textit{English-centric LLMs}}\\ \midrule
\textbf{Gemma 9B-Instruct}  &  67.17  &  66.36  &  67.17  &  65.32  &  67.28  &  66.71  &  59.08  &  59.54  &  59.08  &  58.27  &  58.61  &  59.31  \\

\textbf{Mixtral-8x22B-Instruct}  &  67.86  &  62.54  &  59.42  &  53.87  &  41.93  &  32.14  &  51.21  &  47.40  &  42.43  &  35.72  &  28.79  &  22.31  \\

\textbf{Llama-3.2-90B-Instruct}  &  66.54  &  60.70  &  61.09  &  59.14  &  59.92  &  59.53  &  48.63  &  48.09  &  50.06  &  46.47  &  46.13  &  45.55  \\

\textbf{GPT-3.5}  &  49.59  &  49.94  &  50.29  &  52.95  &  50.52  &  50.75  &  53.53  &  56.65  &  55.72  &  55.38  &  55.14  &  53.53  \\

\textbf{GPT-4o}  &  74.10  &  76.42  &  74.91  &  75.95  &  77.34  &  76.18  &  68.79  &  67.28  &  68.90  &  69.13  &  68.32  &  69.48  \\

\textbf{Claude-3.5-Sonnet}  &  48.72  &  55.60  &  57.49  &  60.86  &  80.99  &  61.94  &  41.16  &  42.89  &  44.97  &  45.78  &  49.25  &  50.29  \\

\textbf{Gemini-Pro-1.5}  &  69.83  &  69.02  &  70.98  &  72.14  &  72.14  &  71.68  &  60.46  &  62.20  &  63.12  &  63.70  &  63.35  &  63.93  \\

\midrule
\multicolumn{13}{c}{\textit{Chinese-centric LLMs}}
\\

\midrule

\textbf{Qwen2.5-0.5B-Instruct}  &  9.94  &  3.47  &  2.31  &  0.46  &  0.34  &  0.23  &  10.75  &  3.70  &  1.04  &  0.81  &  0.58  &  0.58  \\
\textbf{Qwen2.5-1.5B-Instruct}  &  47.63  &  34.11  &  29.48  &  28.90  &  27.28  &  21.62  &  27.75  &  18.96  &  16.30  &  13.41  &  12.60  &  10.06  \\

\textbf{Qwen2.5-7B-Instruct}  &  61.27  &  55.61  &  52.02  &  50.34  &  47.63  &  45.20  &  52.60  &  33.29  &  28.79  &  23.47  &  20.58  &  21.62  \\

\textbf{Qwen2.5-14B-Instruct}  &  71.45  &  72.60  &  71.79  &  71.56  &  72.75  &  71.10  &  65.66  &  63.82  &  65.20  &  65.66  &  66.36  &  65.43  \\

\textbf{Qwen2.5-32B-Instruct}  &  71.79  &  70.84  &  67.71  &  68.80  &  68.66  &  69.75  &  65.43  &  64.39  &  62.66  &  62.89  &  61.73  &  61.16  \\

\textbf{Qwen2.5-72B-Instruct}  &  74.57  &  74.68  &  72.72  &  73.53  &  73.06  &  71.56  &  64.05  &  61.73  &  63.93  &  62.89  &  64.74  &  64.39  \\
 \bottomrule
\end{tabular}
}
\caption{A comparison model performance with Table~\ref{tab: dynamicResult} where the patient agent is played by deepseek-chat model.}
\label{tab: compare_dynamic}
\end{table*}

\section{Robustness to Different Patient Simulators}
~\label{apdx:robustness_simulator}

We further compared two representative Patient simulators, GPT-4o and Deepseek-Chat, to examine whether the OR-dynamic evaluation is sensitive to the choice of the simulated patient model. These two models were selected because they differ substantially in model family and deployment style, providing a useful stress test for the evaluation framework. As shown in Table~\ref{tab: compare_dynamic}, although some absolute accuracy values vary when Deepseek-Chat is used instead of GPT-4o, the overall ranking and performance trends of the Doctor models remain broadly consistent across the two settings. This suggests that the measured capability of the OR Doctor Agent is not dominated by a particular Patient simulator, but instead reflects stable task performance. We ultimately adopted GPT-4o as the default Patient Agent because it produced more consistently controlled and stable patient responses in multi-turn interactions, making it a more suitable simulator for the main experiments, while the comparison with Deepseek-Chat serves as additional evidence for the robustness of the evaluation framework.
\section{Expert Assessment of Simulator Fidelity}
\label{apdx:dilog_N_analss}
To assess the fidelity of the simulated doctor--patient interaction, we conducted a case-based expert annotation study on the generated dialogue trajectories. Specifically, one licensed physician with clinical experience was invited to review each dialogue turn by turn, comparing the simulator's responses against the predefined structured case information. For each turn, the physician examined whether the simulated doctor asked clinically appropriate and coherent follow-up questions, and whether the simulated patient responded in a way that was faithful to the chief complaint, symptom progression, medical history, and family history, without introducing unsupported details. The annotation also considered whether the responses were natural and consistent with realistic outpatient communication. As illustrated in Figure~\ref{fig:sim_dialogue_analysis_cards}, the expert feedback shows that the simulated patient generally preserved key clinical facts accurately across turns, while the simulated doctor asked relevant questions that aligned well with the consultation flow. The annotations also revealed a few minor issues, such as occasional overly structured wording or failure to proactively mention certain salient details, but these issues did not undermine the overall case consistency. Overall, the results provide qualitative evidence that both the simulated doctor and the simulated patient maintain high fidelity to the underlying clinical scenario, supporting the reliability of the simulation framework for downstream evaluation.
\begin{figure*}[t]
\centering
\begin{CJK*}{UTF8}{gbsn}

\turnpair{0}{
\textbf{Simulated Doctor:} Hello, how can I help you?\\
\textbf{模拟医生：} 您好，请问有什么可以帮您？\\[0.25em]
\textbf{Simulated Patient:} For the past two weeks, I feel chest tightness and palpitations when I walk more.\\
\textbf{模拟病人：} 这两周走多一点就会胸闷，心跳有点快。
}{
Accurately reflects the chief complaint (duration + chest tightness + palpitations); no fabricated symptoms. The phrase ``for the past two weeks'' is slightly structured.\\[0.35em]
准确对应主诉（时间 + 胸闷 + 心悸），无虚构症状；“两周”表述略偏结构化。
}

\turnpair{1}{
\textbf{Simulated Doctor:} Does it happen all the time or after activity?\\
\textbf{模拟医生：} 是一直这样还是活动后才出现？\\[0.25em]
\textbf{Simulated Patient:} Mostly after activity like climbing stairs, and it improves after rest.\\
\textbf{模拟病人：} 主要是活动后，比如爬楼梯，歇一下会好一些。
}{
The simulator does not proactively mention the recent 3-day increase in symptom frequency.\\[0.35em]
未主动补充“近 3 天频率增加”信息。\\[1.6em]
}

\turnpair{2}{
\textbf{Simulated Doctor:} Do you have fever, cough, or chest pain?\\
\textbf{模拟医生：} 有没有发烧、咳嗽或者胸痛？\\[0.25em]
\textbf{Simulated Patient:} No fever or cough, and no obvious chest pain.\\
\textbf{模拟病人：} 没有发烧和咳嗽，也没有明显胸痛。
}{
The wording ``no obvious chest pain'' sounds slightly clinical rather than purely colloquial.\\[0.35em]
“明显胸痛”措辞略偏专业化，不太口语。\\[1.6em]
}

\turnpair{3}{
\textbf{Simulated Doctor:} Do you have hypertension or any heart-related history?\\
\textbf{模拟医生：} 有没有高血压或心脏方面的病史？\\[0.25em]
\textbf{Simulated Patient:} I've had hypertension for about ten years.\\
\textbf{模拟病人：} 我有高血压，大概十年了。
}{
Accurately recalls past medical history.\\[0.35em]
准确回忆既往史。\\[2.35em]
}

\turnpair{4}{
\textbf{Simulated Doctor:} Any family history of heart disease?\\
\textbf{模拟医生：} 家里有人得过心脏病吗？\\[0.25em]
\textbf{Simulated Patient:} Yes, my father had coronary artery disease.\\
\textbf{模拟病人：} 有，我父亲得过冠心病。
}{
Fully consistent with structured family history.\\[0.35em]
与家族史完全一致。\\[2.35em]
}
\end{CJK*}
\caption{Expert annotation of simulator fidelity in a multi-turn dialogue case.}
\label{fig:sim_dialogue_analysis_cards}
\end{figure*}

\section{Information Gain}~\label{apdx:entropy}
Table~\ref{tab:entropy_dynamic} report the evolution of model uncertainty during dynamic outpatient referral from two complementary perspectives.
Table~\ref{tab:entropy_dynamic} quantifies uncertainty using entropy over the hypothesis space, providing a normalized measure of decision concentration across turns.
Together, these two tables characterize how interaction reshapes the referral hypothesis space and support the information gain analysis presented in the main text.

\noindent \textbf{Limited Information Gain through Interaction}~ 
From Figure~\ref{fig: infogain}, models such as Mixtral and DeepSeek-v3 exhibit a clear reduction in entropy across dialogue turns, yet show limited gains in referral accuracy. The weak trends observed in Apollo, Qwen3, and LLaMA-3.2-Instruct indicate that entropy reduction and accuracy improvement are only loosely correlated, reflecting limited effective information gain from interaction.

\begin{takeaway}
Strong LLMs benefit from more interactive utterances, but they do not always improve performance with more turns.
\end{takeaway}

\begin{table*}[htb]
\small
\setlength{\tabcolsep}{4pt}
\begin{tabular}{l|r|rrrrrr|rrrrrr}
\toprule
\multicolumn{1}{l|}{\textbf{Models}} &
\multicolumn{1}{c|}{\textbf{\#Params (B)}} &
\multicolumn{6}{c|}{\textbf{Hospital-1}~(Entropy)} &
\multicolumn{6}{c}{\textbf{Hospital-2}~(Entropy)} \\
\midrule
\textbf{Turn} &  & \textbf{0} & \textbf{1} & \textbf{2} & \textbf{3} & \textbf{4} & \textbf{5}
              & \textbf{0} & \textbf{1} & \textbf{2} & \textbf{3} & \textbf{4} & \textbf{5} \\
\midrule
\rowcolor{gray!10} \multicolumn{14}{c}{\textit{Medical LLMs}}\\ \midrule

\textbf{BianQue-2} & 6.5B
& 3.64 & 3.22 & 3.04 & 3.40 & 1.79 & 2.89
& 3.00 & 2.77 & 2.48 & 1.61 & 0.00 & 0.00 \\

\textbf{DISC-MedLLM} & 13B
& 3.69 & 3.58 & 3.22 & 3.26 & 3.00 & 3.26
& 2.71 & 2.56 & 2.48 & 2.56 & 2.48 & 2.56 \\

\textbf{Apollo} & 72B
& 3.22 & 3.04 & 2.94 & 3.09 & 3.00 & 3.04
& 2.48 & 2.30 & 2.20 & 2.40 & 2.20 & 2.30 \\

\midrule
\rowcolor{gray!10} \multicolumn{14}{c}{\textit{English-centric LLMs}}\\ \midrule

\textbf{Gemma-Instruct} & 9B
& 3.61 & 3.58 & 3.53 & 3.56 & 3.61 & 3.58
& 2.56 & 2.48 & 2.40 & 2.48 & 2.56 & 2.48 \\

\textbf{Mixtral-Instruct} & 22B
& 3.71 & 3.64 & 3.26 & 3.30 & 2.71 & 2.77
& 3.00 & 2.94 & 2.89 & 2.94 & 2.83 & 2.89 \\

\textbf{Llama-3.2-Instruct} & 90B
& 3.50 & 3.00 & 3.33 & 2.94 & 2.48 & 2.30
& 3.04 & 2.71 & 2.30 & 2.20 & 2.40 & 2.30 \\

\textbf{GPT-3.5} & 175B
& 3.64 & 3.71 & 3.22 & 2.77 & 2.64 & 2.71
& 2.94 & 2.89 & 2.40 & 2.08 & 1.79 & 1.39 \\

\textbf{GPT-4o} & -
& 3.58 & 3.33 & 3.09 & 2.89 & 2.77 & 1.61
& 2.89 & 2.71 & 2.71 & 2.20 & 2.08 & 1.79 \\

\textbf{o1-mini} & -
& 3.61 & 3.37 & 3.14 & 2.56 & 2.08 & 1.95
& 2.94 & 2.20 & 1.79 & 1.61 & 1.10 & 1.10 \\

\textbf{o1} & -
& 3.56 & 2.83 & 2.40 & 1.95 & 1.61 & 1.39
& 2.48 & 2.08 & 1.61 & 1.10 & 0.69 & 1.39 \\

\textbf{GPT-5} & -
& 3.47 & 3.18 & 3.04 & 2.77 & 1.61 & 1.61
& 2.77 & 1.95 & 1.39 & 1.10 & 0.69 & 1.10 \\

\textbf{Claude-3.5-Sonnet} & -
& 3.61 & 3.33 & 3.09 & 2.94 & 2.83 & 2.77
& 3.14 & 2.89 & 2.71 & 2.64 & 2.56 & 2.56 \\

\textbf{Gemini-Pro-1.5} & -
& 3.64 & 3.40 & 3.18 & 3.40 & 2.89 & 2.83
& 3.14 & 2.94 & 2.77 & 2.64 & 2.56 & 2.56 \\

\midrule
\rowcolor{gray!10} \multicolumn{14}{c}{\textit{Chinese-centric LLMs}}\\ \midrule

\textbf{Yi-1.5-Chat} & 34B
& 3.71 & 3.66 & 3.40 & 3.33 & 3.43 & 2.94
& 3.04 & 3.00 & 3.00 & 2.94 & 3.04 & 3.00 \\

\textbf{DeepSeek V2.5-Chat} & 236B
& 2.94 & 2.83 & 2.71 & 2.77 & 2.64 & 2.71
& 2.64 & 2.20 & 2.08 & 2.48 & 2.08 & 1.61 \\

\textbf{DeepSeek-r1} & 671B
& 3.71 & 3.37 & 3.50 & 2.94 & 2.83 & 2.20
& 3.14 & 2.94 & 2.77 & 2.30 & 1.39 & 1.61 \\

\textbf{DeepSeek-v3} & 671B
& 3.58 & 3.64 & 3.47 & 3.33 & 3.30 & 3.04
& 3.04 & 2.94 & 2.77 & 2.64 & 2.64 & 2.56 \\

\textbf{Moonshot-v1-128k} & -
& 3.00 & 2.94 & 2.83 & 2.89 & 2.77 & 2.83
& 2.30 & 2.30 & 2.20 & 2.20 & 2.30 & 2.20 \\

\textbf{Qwen2.5-Instruct} & 72B
& 3.74 & 3.61 & 3.26 & 3.22 & 3.26 & 3.37
& 2.30 & 2.64 & 2.08 & 2.08 & 2.20 & 1.95 \\

\textbf{Qwen3} & -
& 3.56 & 3.33 & 3.09 & 2.30 & 1.95 & 1.79
& 2.48 & 2.08 & 1.79 & 1.61 & 1.39 & 1.39 \\

\textbf{Kimi-K2} & -
& 3.53 & 3.30 & 2.48 & 2.30 & 2.08 & 2.08
& 3.09 & 2.94 & 1.95 & 1.79 & 2.48 & 1.61 \\

\bottomrule
\end{tabular}
\caption{Entropy computed over the reduced candidate department set at each dialogue turn. Lower entropy indicates a more concentrated referral hypothesis space.}
\label{tab:entropy_dynamic}
\end{table*}

\begin{figure*}[htbp]
\centerline{\includegraphics[width=\textwidth]{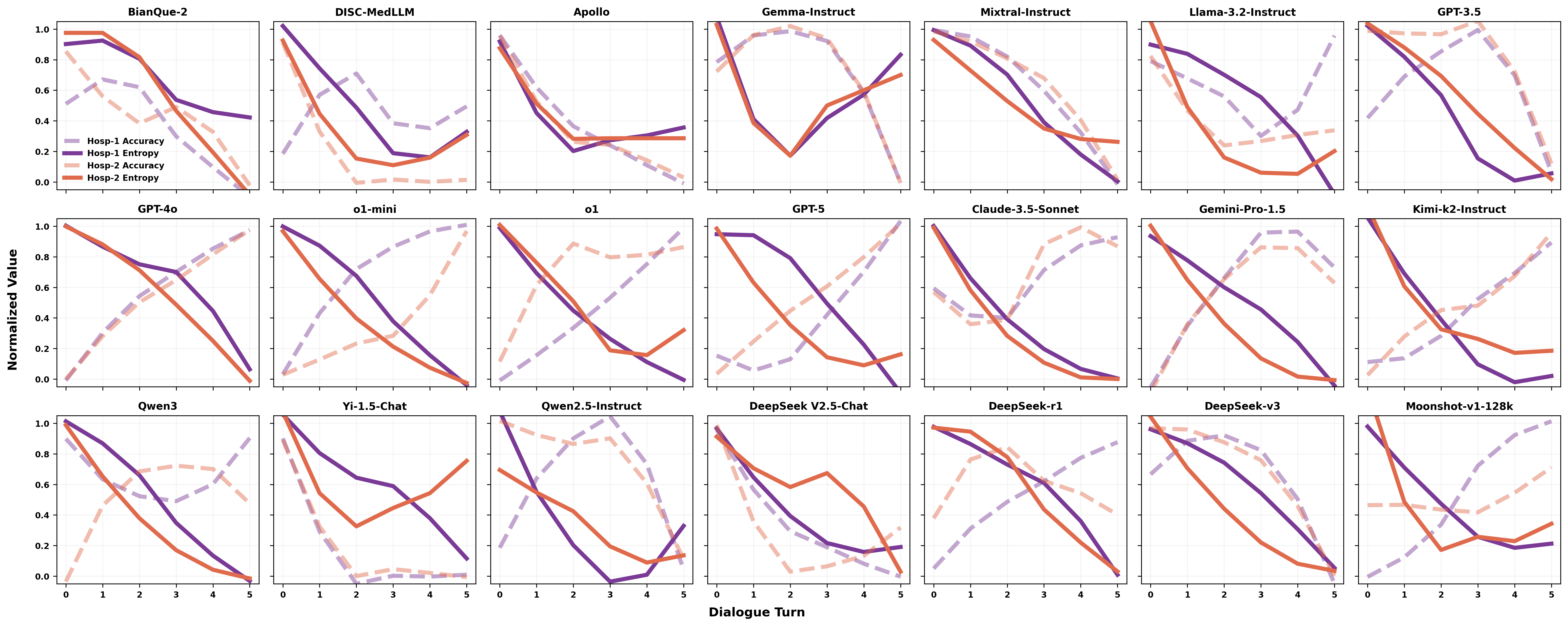}}
\caption{Comparison of normalized accuracy and entropy trajectories for all models, with values min–max normalized to [0, 1] per model and smoothed using a Savitzky–Golay filter~\cite{savitzky1964smoothing}.} 
\label{fig:ig_all}
\end{figure*}

\section{Scaling Law Experiment Design and Interpretation}
\label{apdx:scaling-law-details}

\noindent \textbf{Motivation and Significance}~

While scaling laws are well-known in language modeling, their behavior in complex clinical decision-making tasks, especially involving dynamic multi-turn reasoning, remains underexplored. Our findings indicate that task complexity shapes the saturation threshold: static classification benefits plateau at medium scales (~32B), while dynamic tasks show continued gains with scale, albeit with diminishing returns. This differential behavior suggests that scaling is task-sensitive, and blind reliance on larger models may not yield proportional benefits in all scenarios. The analysis thus provides practical insights into resource allocation and model selection for outpatient triage systems.

\noindent \textbf{Experimental Setup}~

To investigate how model performance scales with parameter size, we conducted a controlled experiment under both static and dynamic outpatient referral scenarios. All experiments are based on the evaluation protocols introduced in Section~\ref{sec:static-evaluation-setup} and Section~\ref{sec:dynamic-evaluation-setup}.

We evaluate a series of Qwen 2.5 models across different sizes (0.5B to 72B), all accessed via API without any additional fine-tuning. In the static scenario, we apply four prompting strategies (zero-shot, few-shot, CoT, and majority vote) to assess data efficiency. In the dynamic scenario, we simulate referral dialogues in two different hospital settings using the same LLMs.

The evaluation metrics and dialogue task structures are consistent with those used in the main experiments. For each setting, performance is averaged over multiple runs or prompts when applicable. Figure~\ref{fig:scaling_law} presents the results.

\noindent \textbf{Key Observations}~

\begin{itemize}
  \item \textbf{Saturation of Gains.} In both static and dynamic settings, model performance improves with size, but gains diminish after 32B. As seen in both subfigures, the performance curve flattens beyond 32B, indicating diminishing returns even as parameter count increases.

  \item \textbf{Prompt Sensitivity in Static Tasks.} In the left plot, smaller models (e.g., 0.5B–3B) show wide performance variance across prompt strategies (e.g., CoT vs. zero-shot), reflected in the shaded standard deviation band. This variation narrows as model size increase, suggesting that larger models are more robust to prompt formats.

  \item \textbf{Difficulty Gap in Dynamic Tasks.} Figure~\ref{fig:scaling_law}, although larger models consistently perform better on both hospitals, the gap between Hospital-1 (complex, 45 departments) and Hospital-2 (simpler, 25 departments) remains largely unchanged across all model sizes. This implies that scaling improves performance, but cannot fully mitigate task complexity.
  \item \textbf{Cross-turn Stability} In the dynamic scenario, performance becomes markedly more stable as model size increases, with 14B--72B models showing much narrower interquartile ranges than 0.5B--7B models in both Hospital-1 and Hospital-2.
  \item \textbf{Cross-task Robustness}This stability gain is consistent across both hospitals, suggesting that scaling improves not only average dynamic accuracy but also the robustness of multi-turn referral decisions under different clinical distributions.
\end{itemize}

\begin{figure*}[htbp]
\centerline{\includegraphics[width=0.8\textwidth]{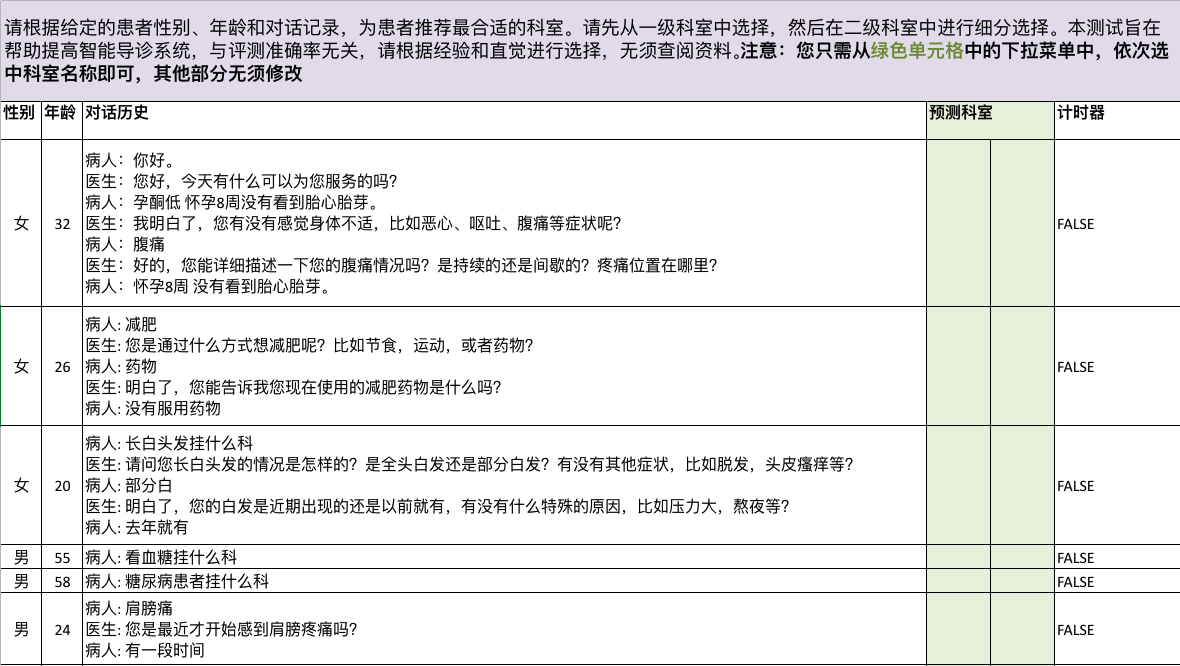}}
\caption{Human evaluation UI for OR-Static Evaluation Task (in Chinese)} 
\label{fig:static-human-eval-cn}
\end{figure*}

\begin{figure*}[htbp]
\centerline{\includegraphics[width=0.8\textwidth]{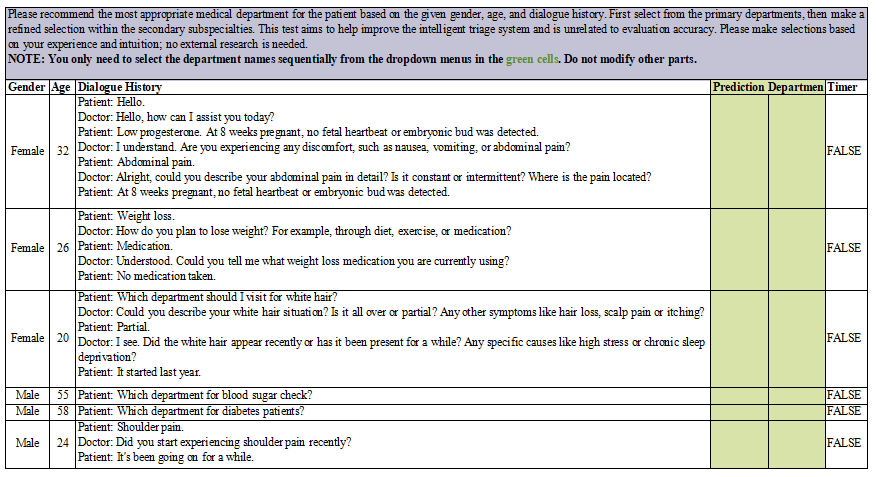}}
\caption{Human evaluation UI for OR-Static Evaluation Task  (in English)} 
\label{fig:static-human-eval-en}
\end{figure*}

\section{Human Evaluation}~\label{apdx: human_eval}
To compare performance trends between humans and LLMs under identical conditions, we invited six human evaluators — two chief physicians, two senior nurses, and two laypersons without medical background — to perform both static and dynamic outpatient referral tasks.
All participants received a brief training on outpatient referral before the evaluation.
In the \textit{static} setting, participants predicted the department based on dialogue histories from 31 randomly sampled cases (Appendix~\ref{apdx: static-human-eval}).
In the \textit{dynamic} setting, participants interacted with LLM-simulated patients, asking follow-up questions as needed and making a referral decisions, using a randomly selected subset of 30 cases (Appendix~\ref{apdx: dynamic-human-eval}).

\subsection{Static Evaluation}~\label{apdx: static-human-eval}
For the static evaluation task, we randomly selected 31 medical dialogue records from the dataset and asked human participants to read the dialogues and recommend the most appropriate department based on the content. The participants were grouped into doctors, nursing staff, and laypersons. After their recommendations, the accuracy for each individual was calculated, as shown in the table. See Figure~\ref{fig:static-human-eval-cn} and Figure~\ref{fig:static-human-eval-en} for the UI for the human evaluation in the OR-Static human evaluation task.

The results in Table~\ref{tab: staticEvaluation} show clear distinctions in performance among the different groups. Both doctors achieved the same accuracy of 85.1\%, indicating consistent decision-making based on their medical expertise. The nursing staff performed similarly, with Nurse-1 matching the doctors' accuracy and Nurse-2 slightly outperforming all others with 89.36\%, which suggests higher familiarity or proficiency in interpreting the dialogues. In contrast, the laypersons scored lower, with Layperson-1 achieving 63.8\% and Layperson-2 at 76.6\%, highlighting the gap in medical knowledge.

\subsection{Dynamic Evaluation}\label{apdx: dynamic-human-eval}
\begin{figure*}[htbp]
\centerline{\includegraphics[width=1\textwidth]{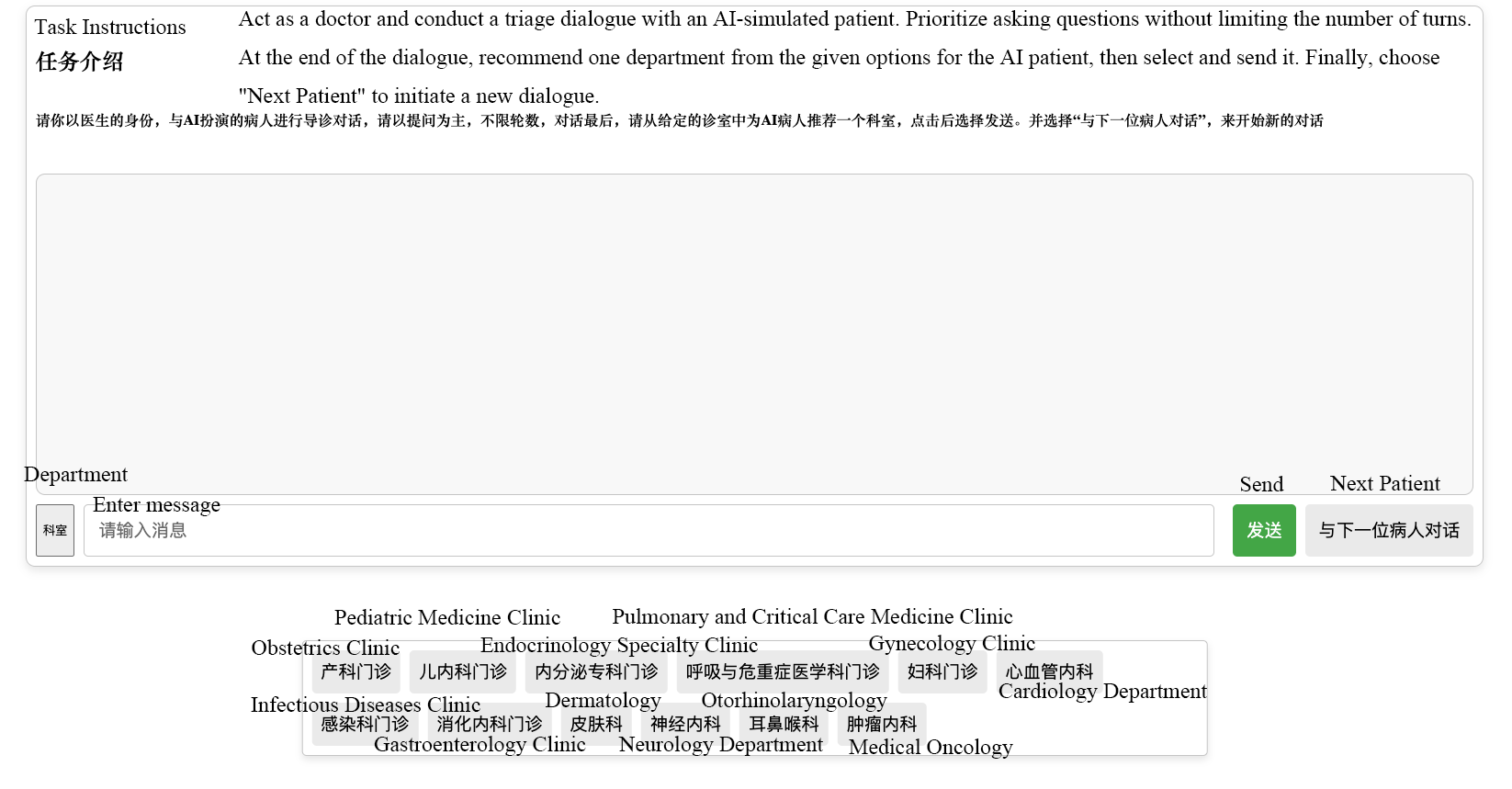}}
\caption{Human evaluation UI for OR-Dynamic Evaluation Task 
} 
\label{fig:dynamic-human-eval}
\end{figure*}

In this task, human participants acted as referral doctors, conversing with an AI-simulated patient. They asked questions to gather information and recommend the appropriate department. This process was repeated for 30 dialogues. See Figure~\ref{fig:dynamic-human-eval} for the UI.

The dynamic evaluation results show clear differences in performance across groups. Doctors performed the best, with doctor-1 achieving 86.67\% accuracy in 1.25 minutes and 6.16 turns, while doctor-2 had 77.42\% accuracy but took longer (6.65 minutes). Nurses showed mixed results, with Nurse-1 having lower accuracy (64\%) but a short duration (1.08 minutes), whereas Nurse-2 had higher accuracy (74.19\%) with more time (4.30 minutes). Laypersons performed significantly worse, with layperson-1 achieving 32.26\% accuracy in 0.50 minutes, and layperson-2 slightly better at 41.94\%, taking longer (4.05 minutes). Overall, medical professionals outperformed laypersons, and there was a trade-off between time spent and accuracy, with longer duration generally leading to higher accuracy.

\subsection{Definition of Accumulated Accuracy.} \label{apdx:acc-clarification}
We define \textit{accumulated accuracy} up to the $t$-th case as the average accuracy of the model over the first $t$ samples:

\begin{equation}
    A_t = \frac{1}{t} \sum_{i=1}^{t} \mathbb{1}_{\mathrm{correct}_i},
\end{equation}

where $\mathbb{1}_{\mathrm{correct}_i}$ equals 1 if the model makes a correct prediction on the $i$-th case, and 0 otherwise.

\subsection{Clarification on Accumulated Accuracy Visualization}\label{apdx:acc-vis}
In Figure~\ref{fig:cumulative_accuracy}, we plot the accumulated accuracy trends for different participant groups in the dynamic evaluation task. The value such as 0.1 shown for the second sample is \textbf{not} a raw accuracy value. To smooth the curve and improve readability, we apply a simple \textbf{local averaging} strategy: instead of plotting the true accumulated accuracy $A_t = \frac{1}{t} \sum_{i=1}^{t} \mathbb{1}_{\mathrm{correct}_i}$, we display the average over a short sliding window (e.g., samples 1 to 3). For example, if only one of the three predictions is correct, the displayed value becomes 0.33.

This smoothed value is used solely for visualization and does \textbf{not} affect the actual evaluation results. Such smoothing is particularly useful when the sample size is small and accuracy varies significantly across samples. A clarification note will be added in the final version of the figure caption to avoid confusion.

From the trend, we observe that some human participants experience declining accuracy as the number of cases increases, likely due to fatigue or external distractions. For instance, doctor-2 and nurse-1 show notable performance drops. In contrast, LLMs maintain stable, time-invariant accuracy throughout the task, underscoring their advantage in handling repetitive medical OR cases.

\end{document}